\documentclass[runningheads]{llncs}
\usepackage{graphicx}
\usepackage{comment}
\usepackage{amsmath,amssymb} 
\usepackage{color}

\usepackage{subfigure}
\usepackage{xcolor}

\begin{document}
	\pagestyle{headings}
	\mainmatter
	\def\ECCVSubNumber{3801}  
	
	\title{Dynamic and Static Context-aware LSTM for Multi-agent Motion Prediction} 

	\titlerunning{Dynamic and Static Context-aware LSTM for Multi-agent Motion Prediction}
	%
	\author{Chaofan Tao\inst{1,2}\orcidID{0000-0002-6093-0854} \and
		Qinhong Jiang\inst{3}\orcidID{0000-0002-5509-7247} \and
		Lixin Duan\inst{2}\orcidID{0000-0002-0723-4016}\and
		Ping Luo\inst{1}\orcidID{0000-0002-6685-7950}}
	\authorrunning{C. Tao, Q. Jiang, L. Duan, P. Luo}
	%
	\institute{
		The University of Hong Kong, Hong Kong, China \and
		University of Electronic Science and Technology of China, China \\
		\email{\{tcftrees,lxduan\}@gmail.com}\\
		\email{\{pluo\}@cs.hku.hk}\\
		\and
		SenseTime, China\\
		\email{\{jiangqinhong\}@sensetime.com}\\	
	}

	
	\maketitle
	
	\begin{abstract}
		Multi-agent motion prediction is challenging because it aims to  foresee the future trajectories of multiple agents (\textit{e.g.} pedestrians) simultaneously in a complicated scene.
		Existing work addressed this challenge by either learning social spatial  interactions represented  by the  positions of a group of pedestrians, while ignoring their temporal coherence (\textit{i.e.} dependencies between different long trajectories), or by understanding the complicated scene layout (\textit{e.g.} scene segmentation) to ensure  safe navigation. 
		However, unlike previous work that isolated the spatial interaction, temporal coherence, and scene layout, this paper designs a new mechanism, \textit{i.e.}, Dynamic  and  Static  Context-aware  Motion  Predictor (DSCMP), to integrates these rich information into the long-short-term-memory (LSTM). It has three appealing benefits.
		(1) DSCMP models the dynamic interactions between agents by learning both their spatial positions and temporal coherence, as well as understanding the  contextual scene layout.
		(2) Different from previous LSTM models that predict motions by propagating hidden features frame by frame, limiting the capacity to learn correlations between long trajectories, we carefully design a differentiable queue mechanism in DSCMP, which is able to explicitly memorize and learn the correlations between long trajectories. 
		(3) DSCMP captures the context of scene by inferring latent variable, which enables  multimodal predictions with meaningful semantic scene layout. 
		Extensive experiments show that DSCMP outperforms state-of-the-art methods by large margins, such as 9.05\% and 7.62\% relative improvements on the ETH-UCY and SDD datasets respectively.
		
		\keywords{Motion prediction, Trajectory Forecasting, Social model}
	\end{abstract}

	\section{Introduction}
	Multi-agent motion prediction is an important task for many real-world applications such as self-driving vehicle, traffic surveillance, and autonomous mobile robot. However, it is challenging because it aims at foreseeing the future trajectories of multiple agents such as pedestrians simultaneously in a complicated scene.
	Existing work \cite{alahi2016social,gupta2018social,xu2018encoding,bisagno2018group,sadeghian2019sophie,choi2019looking,zhao2019multi,manh2018scene} that addressed this challenge can be generally partitioned into two categories. In the first category \cite{alahi2016social,gupta2018social,xu2018encoding,bisagno2018group}, previous work predicted the motions by learning social spatial interactions, which are represented by the positions of pedestrians. However, these approaches typically ignored the the dependency between different long trajectories of pedestrians. In the second category \cite{sadeghian2019sophie,zhao2019multi,manh2018scene,choi2019looking}, prior arts combined scene  understanding  to regularize the predicted trajectory, such as visual feature of the complicated  scene  layout.
	
\begin{figure*}[t]
		\centering
		\subfigure[Illustration of multi-agent motion prediction.]{
			\begin{minipage}[t]{0.35\linewidth}
				\centering
				\includegraphics[width=1.6in,height=1.3in]{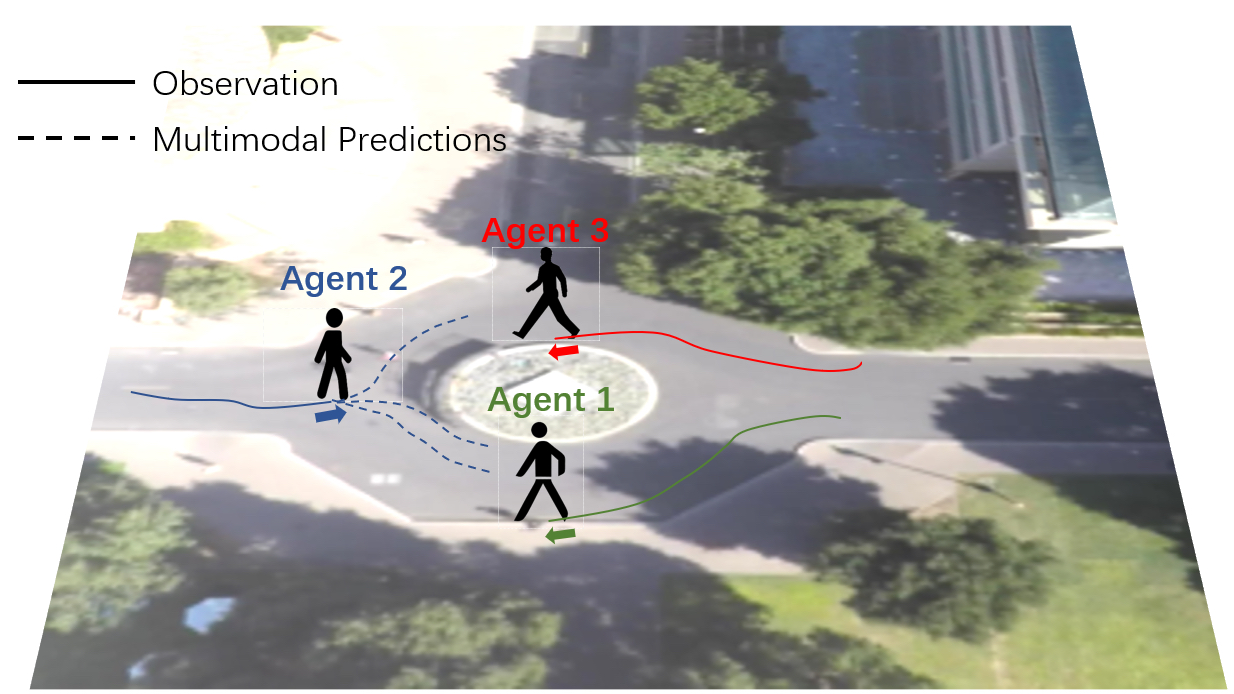}
				\label{figure1a}
			\end{minipage}
		}
		\subfigure[Comparisons of different approaches.]{
			\begin{minipage}[t]{0.57\linewidth}
				\centering
				\includegraphics[width=2.6in,height=1.4in]{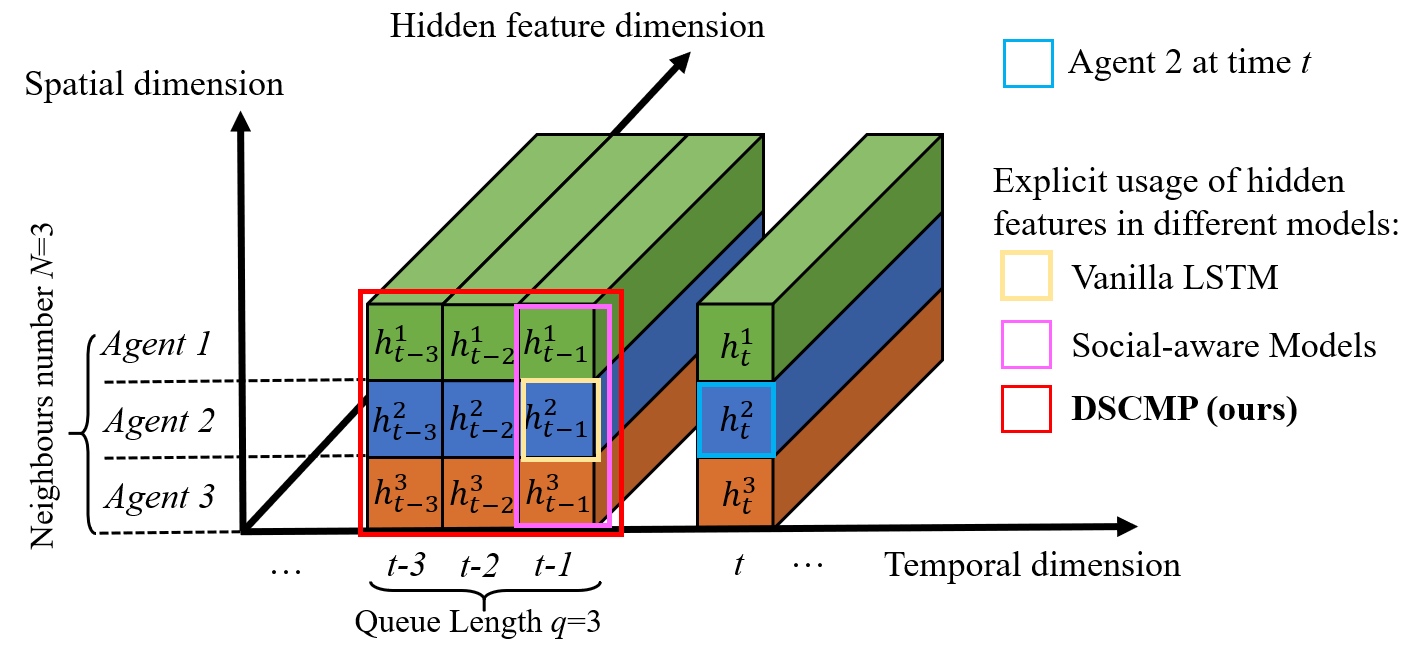}
				\label{figure1b}
			\end{minipage}
		}
		
		\subfigure[Comparisons of the Relative Average Distance Error (ADE) on various datasets.]{
			\label{figure1c}
			\begin{minipage}[t]{0.15\linewidth}
				\centering
				\includegraphics[width=0.8in,height=0.7in]{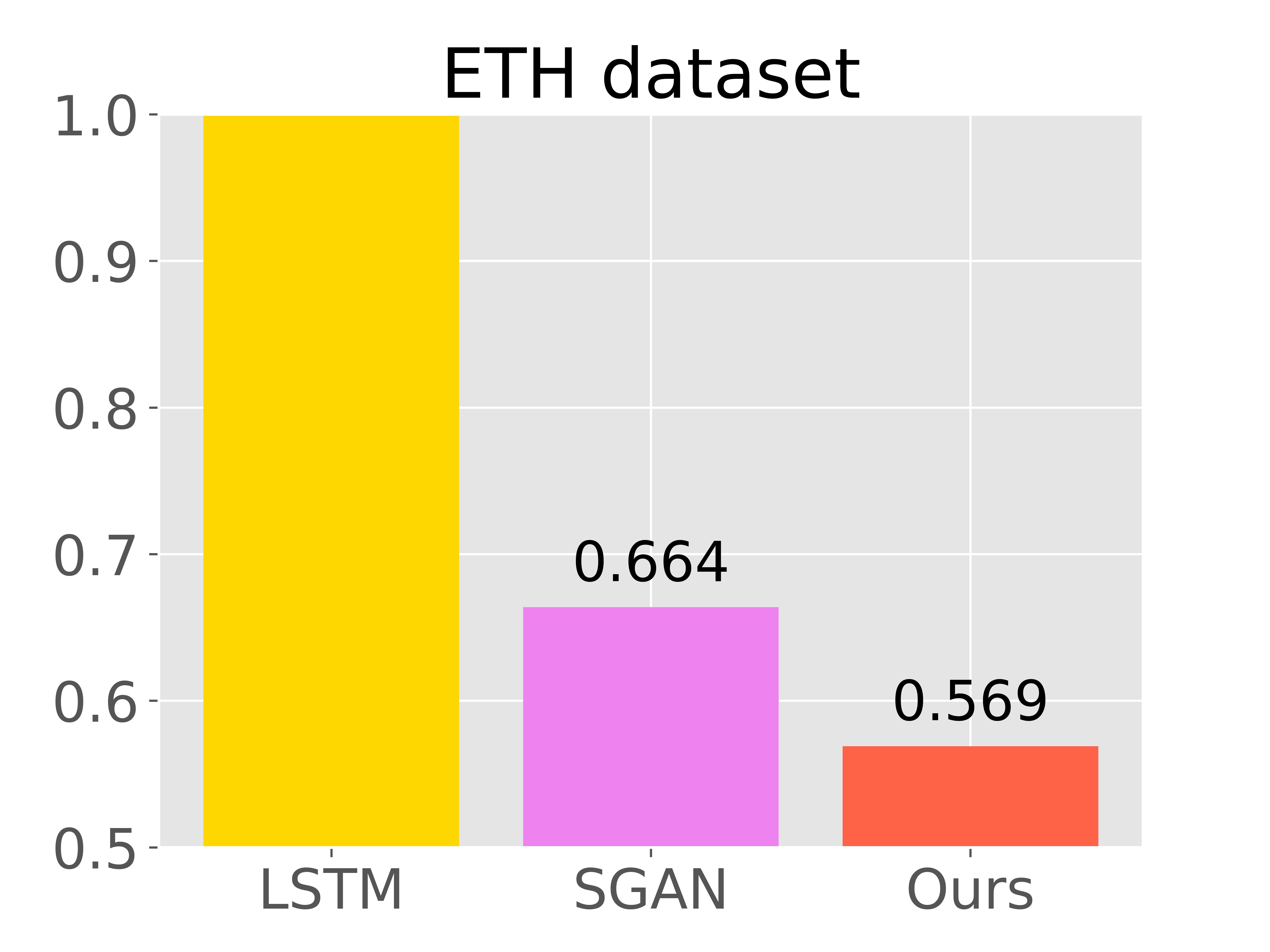}
			\end{minipage}	
			\begin{minipage}[t]{0.15\linewidth}
				\centering
				\includegraphics[width=0.8in,height=0.7in]{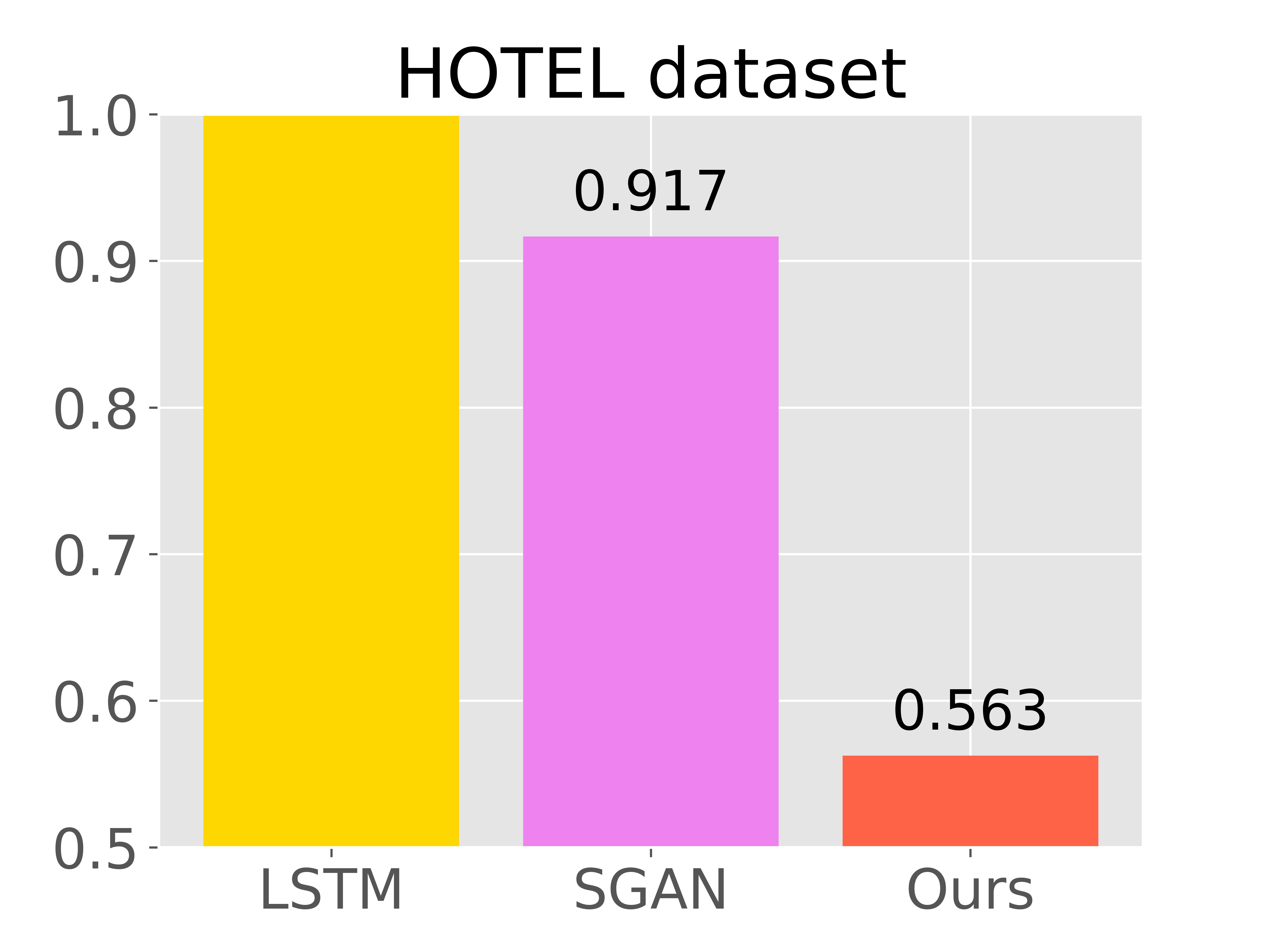}
			\end{minipage}
			\begin{minipage}[t]{0.15\linewidth}
				\centering
				\includegraphics[width=0.8in,height=0.7in]{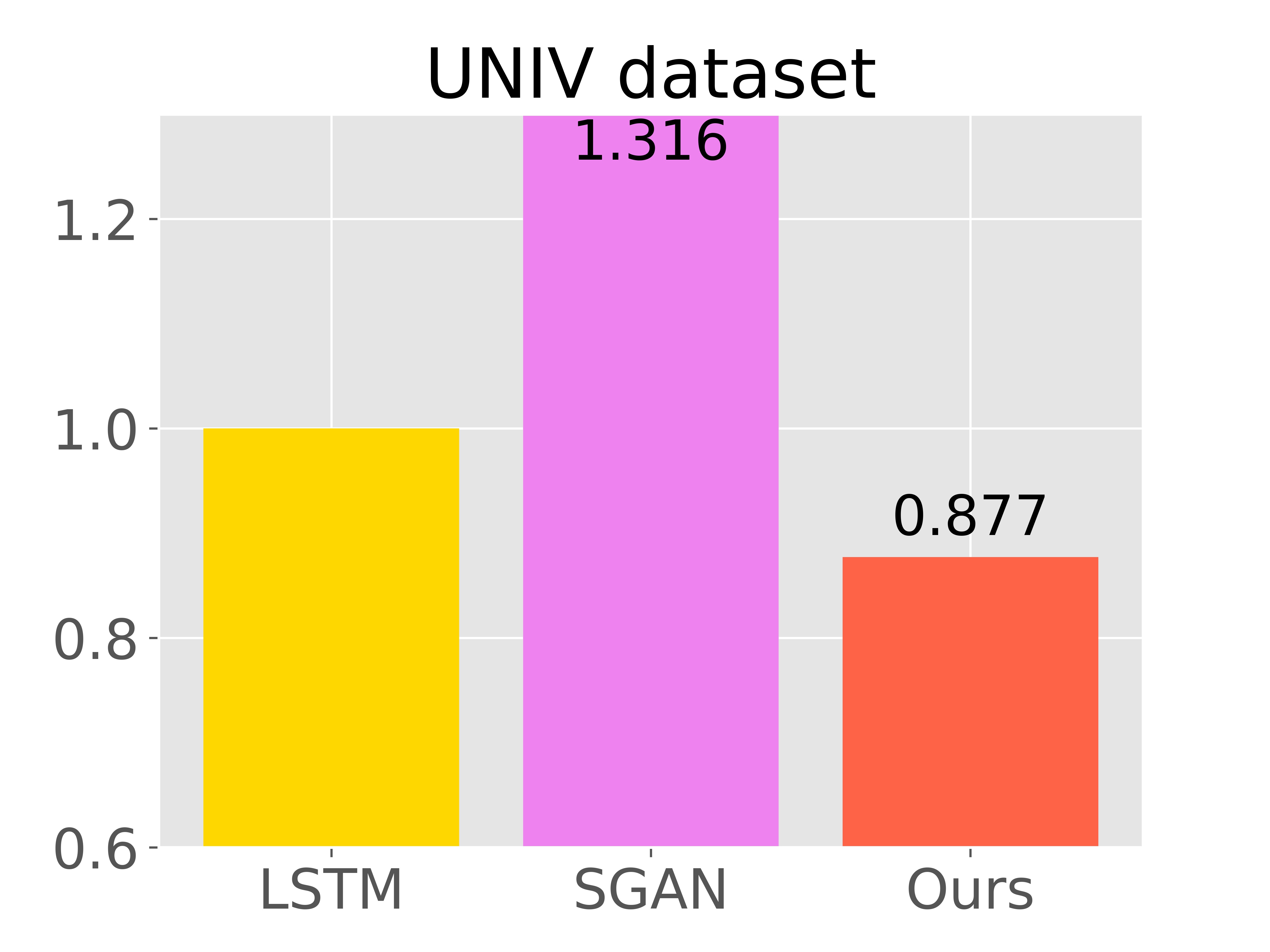}
			\end{minipage}
			\begin{minipage}[t]{0.15\linewidth}
				\centering
				\includegraphics[width=0.8in,height=0.7in]{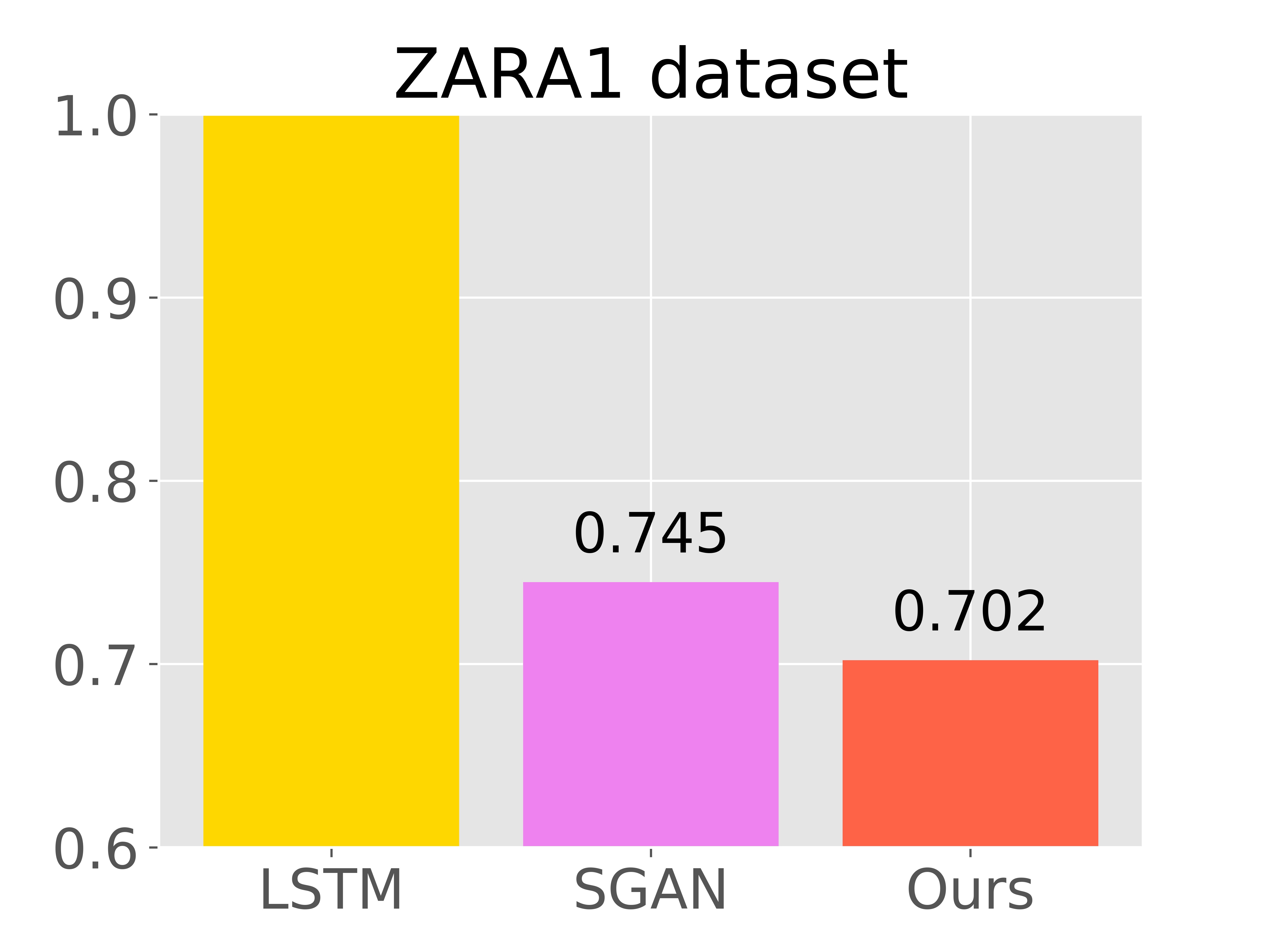}
			\end{minipage}
			\begin{minipage}[t]{0.15\linewidth}
				\centering
				\includegraphics[width=0.8in,height=0.7in]{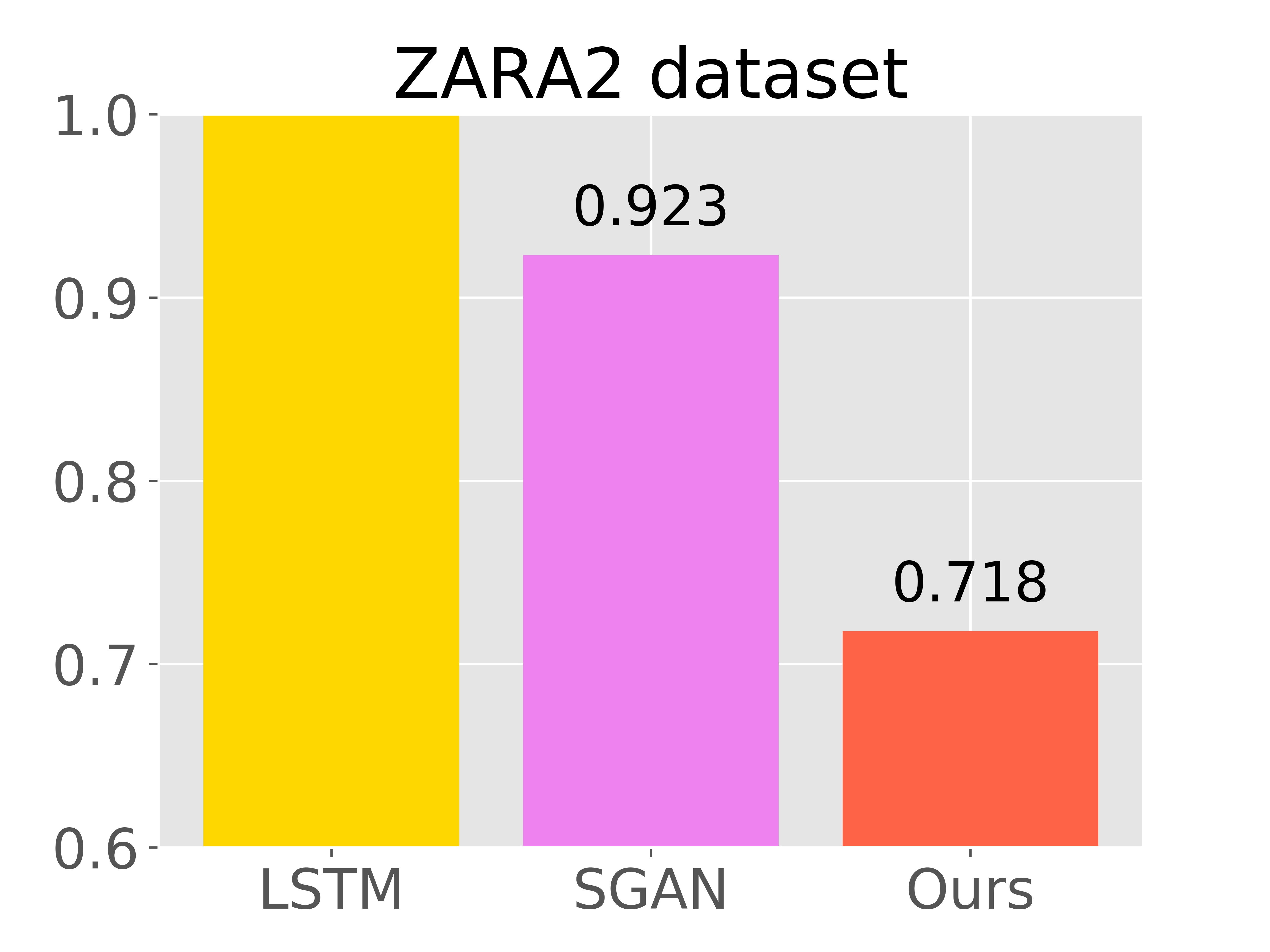}
			\end{minipage}
			\begin{minipage}[t]{0.15\linewidth}
				\centering
				\includegraphics[width=0.8in,height=0.7in]{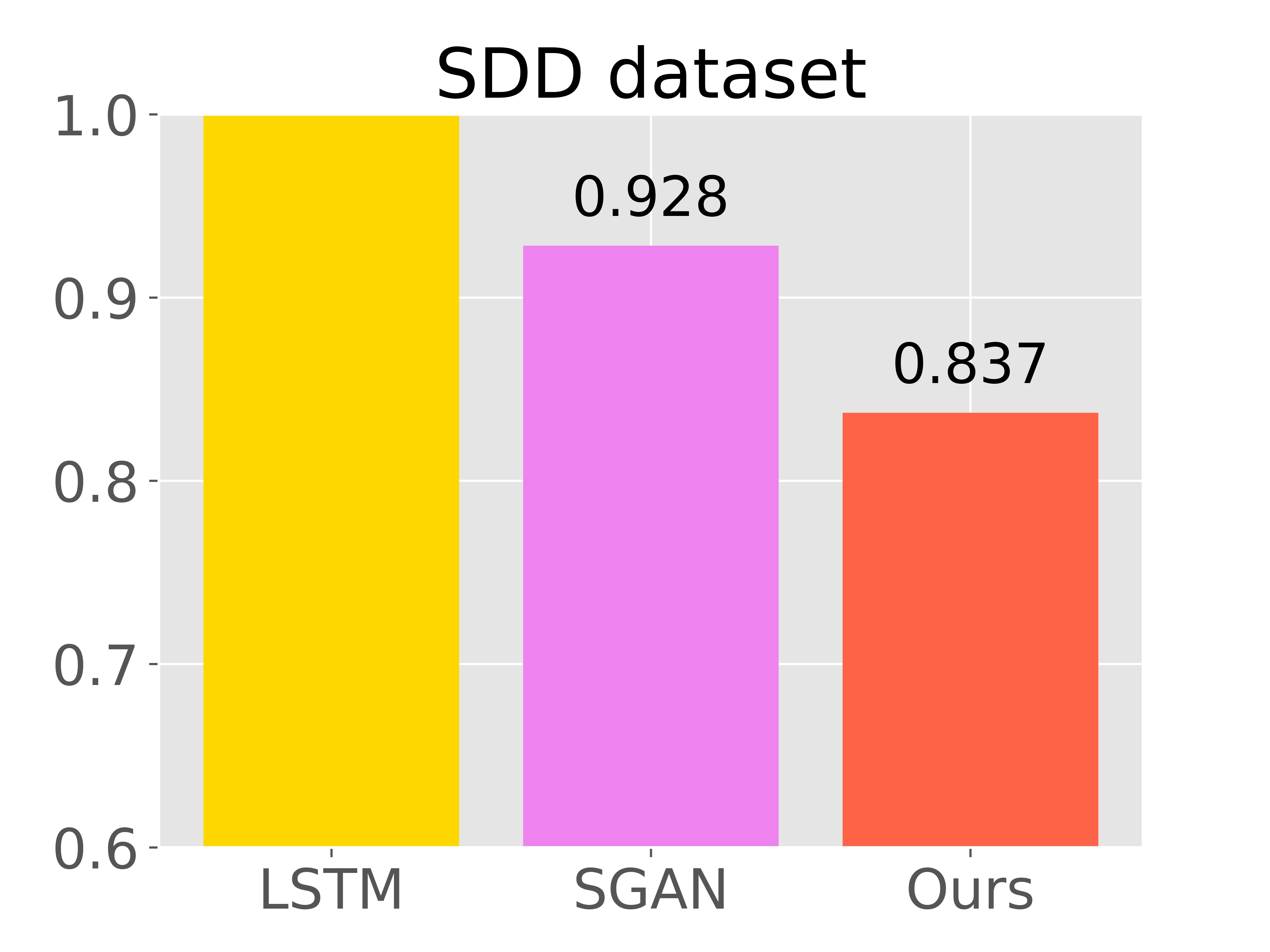}
			\end{minipage}	
		}
		
		\centering
		\caption{ \textbf{(a)} illustrates motion prediction in a real-world scenario, where both the dynamic motion context across  agents and static  context are involved. \textbf{(b)} compares various methods including LSTM, social-aware models (\textit{e.g.} SGAN \cite{gupta2018social}) and DSCMP. The queue mechanism in DSCMP enriches the scope of dynamic context at each frame, enabling DSCMP to effectively learn long motions. \textbf{(c)} compares the Average Distance Error (ADE) of LSTM \cite{hochreiter1997long}, SGAN  \cite{gupta2018social} and DSCMP on  ETH (two subsets: ETH, HOTEL) \cite{pellegrini2009you}, UCY (three subsets: UNIV, ZARA1, ZARA2) \cite{lerner2007crowds}, SDD \cite{robicquet2016learning}  by using the performance of LSTM as baseline (\textit{i.e.} unit 1; lower value is better). We see that DSCMP surpasses its counterparts by large margins. Best viewed in color.}
		\label{figure1}
		\vskip -0.5cm
	\end{figure*}

	Different from existing work that either model agents' interactions or the scene layout, we carefully designed novel mechanisms in LSTM to model dynamic interactions of pedestrians in both spatial and temporal dimensions, as well as modeling the semantic scene layout as latent probabilistic variable to constrain the predictions. These design principles enable our model to predict multiple trajectories for each agent that cohere in time and space with the other agents. We see that the proposed method outperforms its counterparts in many benchmarks as shown in Fig.\ref{figure1c}.

	We name the proposed method as  Dynamic and Static Context-aware Motion Predictor (DSCMP), which has an encoder-decoder structure of LSTM that has carefully devised mechanisms to tackle multi-agent motion prediction.
	DSCMP has three appealing benefits that previous work did not have.
	
	For the first benefit, unlike existing methods \cite{alahi2016social,gupta2018social,xu2018encoding,sadeghian2019sophie,zhao2019multi} that employed recurrent neural networks (RNNs) to learn motion by passing message frame by frame, DSCMP incorporates a queue mechanism in LSTM to explicitly propagate hidden features of multiple frames, enabling to capture long trajectories among pedestrians more explicitly than prior arts.
	
	Specifically, the vanilla LSTM in previous approaches \cite{fragkiadaki2015recurrent,shah2016applying} attempts to learn a frame-by-frame predictor for each agent $i$, denoted as $m_{t+1}^i = p(m_t^i, h_{t-1}^i)$, where $p(\cdot)$ is a prediction function of LSTM, $m_t^i$ represents the current motion state (\textit{i.e.} the $x,y$ positions) at the $t$-th frame and $h_{t-1}^i$ is the hidden feature of the previous single frame. The frame-wise models hinders their capacity to capture the dependency between long trajectories of pedestrians.

	The recent approaches of social-aware LSTM models \cite{alahi2016social,gupta2018social,vemula2018social} modified the above vanilla LSTM by using $m_{t+1}^i = p(m_t^i, \bigcup_i^{N(i)} h_{t-1}^i)$, where $\bigcup$ denotes combination of a set of hidden features of the spatial neighbors of the $i$-th pedestrian (denoted by $N(i)$) at the previous $(t-1)$-th frame.

	However, the above methods are insufficient to consider  the interactions across agents. For example, as shown in  Fig.\ref{figure1a}, the agent $2$ is heading towards agent $1$ and agent $3$. To avoid collision, the agent $2$  tends to adjust his future trajectory by anticipating the intention of agent $1$ and $3$  based on their recent movement history, rather than their states  at the previous one frame only. 

	Different from the above existing approaches, a LSTM is carefully designed in DSCMP to learn both spatial dependencies and temporal coherence of moving pedestrians. The LSTM contains two modules, including an  Individual Context Module (ICM) and a Social-aware Context Module (SCM). As shown in Fig.\ref{figure1b}, our model fully understands spatial and temporal contexts across agents  by learning a predictor denoted as $m_{t+1}^i=p(m_t^i,  \bigcup_i^{N(i)}  Q_t^i)$, where $Q_t^i$ denotes a set of features not only across agents at a certain frame, but also across multiple successive frames of different agents.

	More specific, the ICM of DSCMP passes feature of the current motion state and the corresponding feature queue into LSTM cell. Multiple forget gates control the information flow of the frames in the queue. At each iteration, we update the queues by appending the features of the latest frame, and popping out the earliest features. Furthermore, the SCM of DSCMP refines the updated queues by using the queues of the neighboring agents. Since these queues preserve agent-specific motion cues in the past multiple frames, we are able to learn a long-range spatial-temporal interactions with the aggregation of queues.

	For the second benefit, we observe that  the future movements of agents in real scenarios have uncertainty, since multiple trajectories are plausible.
	For instance, an agent would naturally consider his/her surrounding scene layout when deciding his/her possible future paths. In particular, an agent could turn left or right in a crossroad, while he/she has limited choices around a street corner. However, the recent methods either neglected the guidance of scene layout to produce diverse predictions for each agent, or even totally ignored the scene information. 
	In contrast, DSCMP incorporates the scene information into the learning of diverse predictions by using $m_{t+1}^i=p(m_t^i,  \bigcup_i^{N(i)}  Q_t^i,I)$, where $I$ indicates the semantic scene feature after scene segmentation. In practice, this semantic scene feature is modeled as a latent variable of a probabilistic distribution to predict multiple future trajectories for each agent.
	
	For the third benefit, in order to understand the uniqueness of DSCMP, we propose a new evaluation metric called Temporal Correlation Coefficient (TCC) to fully evaluate the temporal correlation of motion patterns,  bridging the gap where the commonly used metric such as Average  Distance  Error (ADE) and Final Distance Error (FDE) are insufficient to evaluate temporal motion correlations.
	Extensive experiments on  dataset ETH \cite{pellegrini2009you}-UCY \cite{lerner2007crowds}, SDD \cite{robicquet2016learning} show that DSCMP surpasses state-of-the-art methods on all the above metrics by large margins, such  as  9.05\% and 7.62\% relative improvements on metric ADE compared to the latest method STGAT \cite{huang2019stgat} method.
	
	To summarize the above benefits, this work has three main \textbf{contributions}. \textbf{(1)} We present a novel future motion predictor, named DSCMP, which is able to explicitly model both the spatial and temporal interactions between different agents, as well as producing multiple probabilistic predictions of future paths for each agent.   
	\textbf{(2)} We carefully design the LSTM modules in DSCMP to achieve the above purposes, where all modules can be trained end-to-end including the  Individual Context Module (ICM), the Social-aware Context Module (SCM), and a latent scene information module.  
	\textbf{(3)} Extensive experiments on the ETH \cite{pellegrini2009you}, UCY \cite{lerner2007crowds}, and SDD \cite{robicquet2016learning} datasets demonstrate that DSCMP outperforms its counterparts by large margin in multiple evaluation metrics such as ADE and FDE, as well as a new metric, Temporal Correlation Coefficient (TCC), proposed by us to better examine the temporal correlation of motion prediction.
	
	\section{Related Work}
	
	\noindent{\bf Motion Prediction}  Early approaches for motion prediction \cite{rudenko2019human} like physics-based methods\cite{pellegrini2009you,yamaguchi2011you,petrich2013map} and planning-based methods \cite{yi2016pedestrian,vasquez2016novel,lee2017desire,karasev2016intent} are usually limited by hand-crafted kinematic equations and reward function respectively. With the development of recurrent neural networks, pattern-based methods \cite{alahi2016social,gupta2018social,vemula2018social,kosaraju2019social,zhang2019sr,kosaraju2019social,ivanovic2019trajectron,xu2018encoding,park2018sequence,kim2017probabilistic} have been studied recently.  While most of the models consider the agents in world coordinates, some work \cite{park2018sequence,kim2017probabilistic} explore trajectory prediction with egocentric vision.
	
	A pioneering pattern-based work that combine the LSTM and social interactions is introduced in \cite{alahi2016social}. The authors of \cite{gupta2018social} proposes an adversarial framework to sequentially generate predictions. A social pooling is employed to learn the spatial dependencies among agents. Spatio-temporal graphs \cite{vemula2018social,ma2019trafficpredict,ivanovic2019trajectron} are adopted to model the relations on a complete graph, whereas these methods suffer from implicit modelling of dynamic edges or poor scalability with $O(N^2)$ complexity. STGAT \cite{huang2019stgat} is the most relevant method to our work.  It considers the temporal correlations of motion in multiple frames. Unlike STGAT deduces a single  correlation representation from the whole observation process, we explicitly keep track of the temporal correlation for each iteration during observation.  We also take scene context into consideration.
	
	\noindent{\bf Contextual Understanding} Humans are capable of inferring and reasoning by understanding the context. Rich contextual information is proven to be valuable in sequential data modeling (e.g. video, language, speech). Attention mechanism \cite{vaswani2017attention,zhang2018self} has shown great success in concentrating on the significant parts of visual or textual inputs at each time steps. Non-local operation \cite{wang2018non} works as a generic block to capture the dependencies directly by computing the pair-wise relations in the long-range context. In the field of motion prediction, graph attention \cite{huang2019stgat,kosaraju2019social} assigns  different importance to the neighboring pedestrians to involve the social-aware interactions. The authors of \cite{zhao2019multi,sadeghian2019sophie} encode  visual features of scene context in the LSTM to predict physics-feasible trajectories.
	
	\noindent{\bf Multimodal Predictions} Multi-modality is an important characteristic of motion prediction that implies the multiple plausible choices of future trajectories. To model this uncertainty, the model is required to generate diverse predictions. A common approach \cite{gupta2018social,zhao2019multi,sadeghian2019sophie} is to fuse latent variables sampled from a predefined Gaussian distribution $\mathcal{N}(0,1)$ with hidden feature. However, predefined latent variables suffer from the absense of context reasoning. In this paper, the latent variable is learnable from the scene context, which enables our model to generate multimodal predictions with meaningful semantics.

	\section{Method}
	\begin{figure*}[t]
		\centering
		\includegraphics[height=4.9cm]{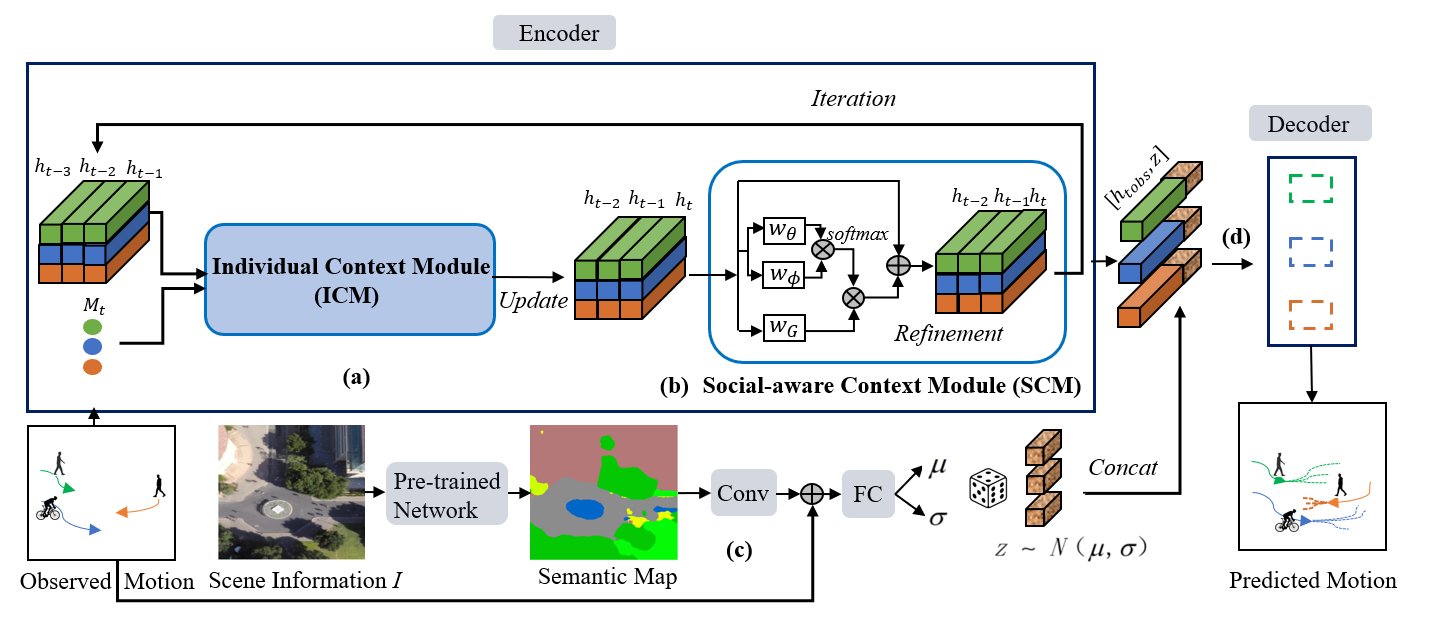}
		\caption{The overview of our framework (DSCMP). Given a sequence of the observed motions, we construct agent-specific queue to store the LSTM features of previous frames within a queue length. The queue length is set to 3 as example. \textbf{(a)} For each iteration during observation, the current motion state and queues are encoded via ICM. \textbf{(b)} The queues are updated by appending the latest features and popping the earliest ones. In the SCM, the queues are adaptively refined by considering the pair-wise relations of features in the neighbours' queues. \textbf{(c)} The semantic map of static context is incorporated with observed motion to generate a learnable latent variable $z$. \textbf{(d)} We concatenate the last hidden feature in the encoder and the latent variable to predict the motions via LSTM decoder.}
		\label{figure2}
		\vskip -0.5cm
	\end{figure*}
	\textbf{Overview.} To be specific, during the observation from frame $t_{1}$ to $t_{obs}$, the motion states of all $N$ agents $M_{t_1:t_{obs}} =  \left \{  M_{t} |  t \in [ t_1, \cdots ,t_{obs} ]  \right \}$ in a scene and scene information $I$  are given, where $M_{t}\in R^{N\times d}$. Symbol $d$ is the dimension of input motion state. It refers to the x-y coordinates of agent's location in this paper, thereby $d=2$.  $ M_{t}  =  \left \{ M_{t}^{i} | i \in N \right \}$,  where $M_{t}^{i}=(x_t^i, y_t^t) \in R^{2}$ denotes the location of agent $i$ at frame $t$ . Our goal is to predict the locations of all the agents $M_{t_{obs+1}:t_{obs+pred}}$ in the future frames. 
	The workflow of our framework DSCMP is illustrated in Fig.\ref{figure2}. For each iteration during observation, we send the current motion states with proposed queues to the encoder, and then we update the queues via  ICM and then refine them via SCM. The last hidden features in the encoder is concatenated with a scene-guided latent variable. Later, the fused feature is passed to a LSTM decoder to obtain predicted motion.
	
	\subsection{The Function of Queues}
	With the setup of queues, the previous motion context and memory context for current frame $t$ are temporarily stored. Specifically, we construct a  hidden feature queue $Q_{h_t}^i= \left [ h_{t-q}^i, \cdots, h_{t-1}^i \right ] \in R^{1*q*h}$  and cell queue $Q_{c_t}^i= \left [ c_{t-q}^i, \cdots, c_{t-1}^i \right ] \in R^{1*q*h}$ for each agent $i$ The size of LSTM feature is denoted as $h$. The queue length $q$ describes a time bucket that the features are explicitly propagated. We initialize the hidden feature queues  and cell queues  with zero for each agent. 
	
	\subsection{Individual Context Module}
\begin{figure*}[t]
		\centering
		\subfigure{
			\begin{minipage}[t]{0.50\linewidth}
				\centering
				\includegraphics[width=2.2in, height=1.35in]{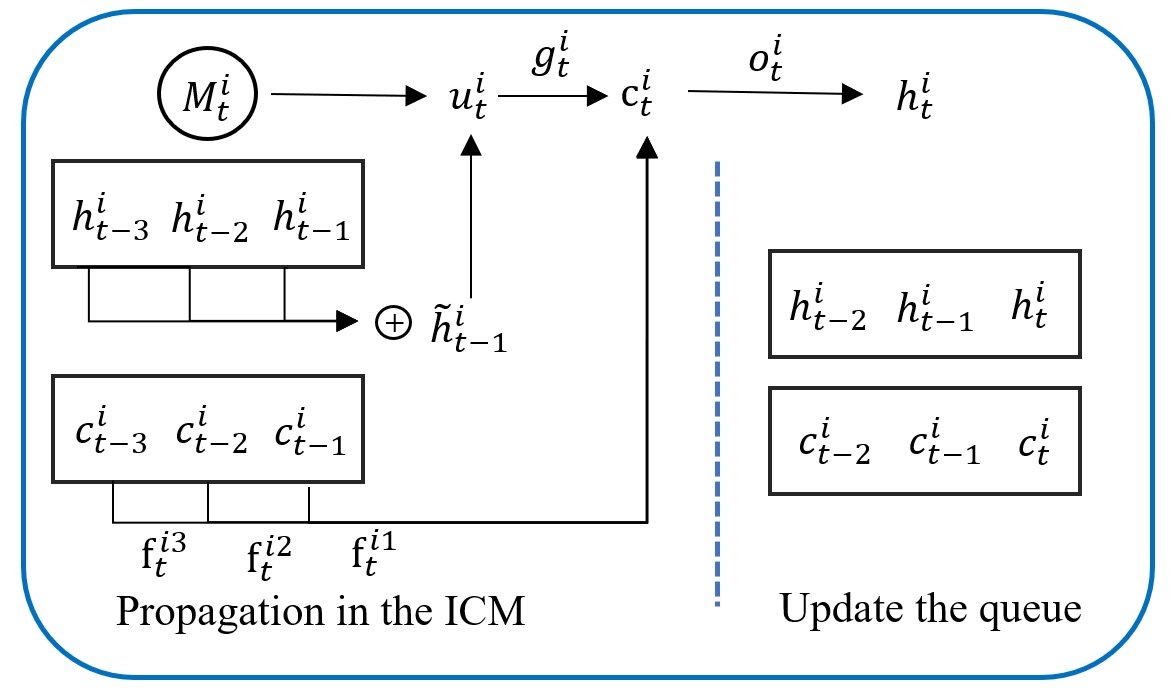}
				\label{figure3a}
			\end{minipage}%
		}
		\subfigure{
			\begin{minipage}[t]{0.44\linewidth}
				\centering
				\includegraphics[width=1.8in, height=1.35in]{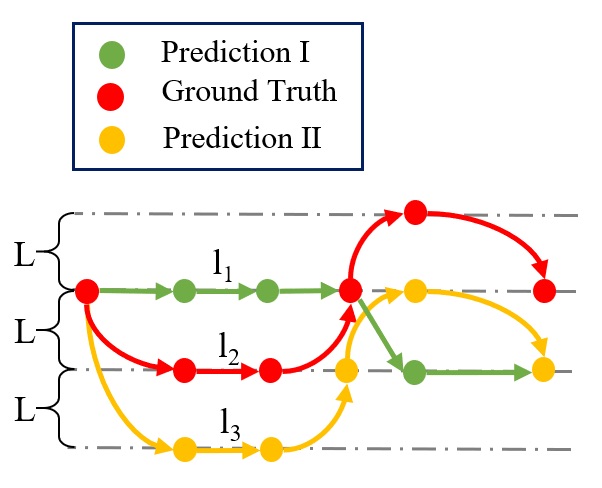}
				\label{figure3b}
			\end{minipage}
		}
		\centering
		\caption{ \textbf{(a):} The computation graph of Individual Context Module (ICM).  \textbf{(b):} An  example to show Euclidean distance-based metric (e.g. ADE and FDE) cannot fully evaluate motion patterns. The prediction \uppercase\expandafter{\romannumeral1} and prediction \uppercase\expandafter{\romannumeral2} score the same ADE (5L/6) and FDE (L), whereas the prediction \uppercase\expandafter{\romannumeral2} captures the temporal correlation of motion pattern much better than the prediction \uppercase\expandafter{\romannumeral1}.}
		\label{figure3}
	\end{figure*}

	Based on the aforementioned queues, we firstly capture the temporal correlations of trajectories from individual level. In order to handle with multiple inputs in one iteration, we employ a tree-like LSTM cell \cite{tai2015improved}. The historical states in a time bucket ($t-q, \cdots, t-1$) are viewed as the children of the current state $t$. After an iteration, the queues are updated by appending features at frame $t$ and popping features at frame $t-q$
	We first average the hidden features in the queue to obtain a holistic representation $\tilde{h}_{t-1}^i=\sum_{l = 1}^{q} h_{t-l}^i$ from the past frames. As the computational graph of ICM shown in Fig.\ref{figure3a}, the propagation is formulated as follows:
	
	\begin{equation}
	\begin{aligned}
	& g_t^i =\sigma (W^g M_t^i + U^g \tilde{h}_{t-1}^i + b^g), & f_t^{il} =\sigma (W^f M_t^i + U^f h_{t-l}^i + b^f),\\
	& o_t^i =\sigma (W^o M_t^i + U^o \tilde{h}_{t-1}^i + b^o), 	& u_t^i =\tanh (W^u M_t^i + U^u \tilde{h}_{t-1}^i + b^u),\\
	& c_t^i = g_t^i \odot u_t^i + \sum_{l=1}^{q} f_t^{il} \odot c_{t-l}^i, & h_t^i = o_t^i \odot \tanh (c_t^i),\\
	\end{aligned}
	\label{eq1}
	\end{equation}
	where $\sigma$ is sigmoid function and $\odot$ is element-wise multiplication. From the equation \ref{eq1}, we could observe that multiple previous frames pass message to the current cell. The contributions of these frames to the current state are controlled by multiple forget gates $f_t^{il}, l \in \left[1,q \right]$. In the case $q=1$, ICM degenerates to a vanilla LSTM cell, which considers the previous single feature only at one iteration.
	
	In practice, we assign each agent the queues of the fixed length. We point out that it is inappropriate in some cases. For example, the motion of some agents may be erratic that temporally incoherent with the past states. However, the adaptive forget gates could control the volume of information from the past frames. Hence, irrelevant motions could be filtered during the propagation.  
	
	\subsection{Social-aware Context Module}
	Social interactions works as an important part of dynamic context.  Since the aggregation of neighbors' queues $\in R^{N(i)*q*h}$ store the surrounding historical information across agents, our model could learn spatio-temporal dependencies in a single operation. Here $N(i)$ denotes the number of neighbors of the agent $i$ (include oneself). In the SCM, we compute the pair-wise relations of elements in the queues from neighbours. The refined queues can be viewed as a weighted sum from the neighbors' queues.  Non-local block \cite{wang2018non} is chosen for relation inference since it not only captures distant relations but also keeps the shape of input. The refinement of the hidden feature queue $Q_{h_{t+1}}^i= \left [ h_{t-q+1}^i, \cdots, h_{t}^i \right ]$ is computed as:
	\begin{equation}
	h_{t-l}^i = h_{t-l}^i  \oplus  \frac{1}{\mathcal{Z}_{t-l}^i} \sum_{j=1}^{N(i)} \mathcal{R}(h_{t-l}^i, h_{t-l}^{j}),  \mathcal{G}(h_{t-l}^{j}) , \ l \in  \left \{ 0, \cdots, q-1  \right \}
	\label{eq8-2}
	\end{equation}
	\begin{equation}
	\mathcal{R}(h_{t-l}^i, h_{t-l}^{j})=(W_{\theta} h_{t-l}^i)^\mathrm{T} (W_{  \phi} h_{t-l}^{j}), \quad  \mathcal{G}(h_{t-l}^{j})=W_{\mathcal{G}} h_{t-l}^{j},
	\label{eq8-3}
	\end{equation}
	where $\mathcal{R}(h_{t-l}^i, h_{t-l}^{j})$ is a scalar that reflects the relationships between the feature $h_{t-l}^i$ and $h_{t-l}^{j}$. And the component $\mathcal{G}(h_{t-l}^{j})$ refers to the transformed representation of the neighbor agent $j$ at frame $t-l$. $N(i)$ denotes the neighbors of agent $i$. $\mathcal{Z}_{t-l}^i=\sum_{j=1}^{N(i)} \mathcal{R}(h_{t-l}^i, h_{t-l}^{j})$ is a normalization factor. The parameters of function $\mathcal{R}(\cdot, \cdot)$ and $\mathcal{G}(\cdot)$ are shared among agents. The cell queues stay invariant since we focus on motion interactions rather than memory at this step.

	\subsection{Semantic Guidance from Scene Context}
	Scene information is a valuable static context that provides the semantic of layout around the agents.  In practice, we  extract the semantic maps from resized $256 \times 256$ scene images $I$ via pre-trained PSPNet \cite{zhou2017scene,zhao2017pyramid} off-line.  After that, we send the semantic maps to  convolutional layers (Conv) and then combine them with the observed trajectories via a  fully-connected (FC) layer. The latent variable $z$ is obtained with reparameterization trick on the mean $\mu$ and variance $\sigma$ as follows:
	\begin{equation}
	[\mu, \sigma]= \textrm{FC}(\textrm{Conv}(I) \oplus \sum_i^{N(i)} M^i_{t_1:t_{obs}}),  \quad
	z \sim  \mathcal{N}(\mu, \sigma),
	\label{eq9}
	\end{equation}
	where $\oplus$ denotes element-wise addition. The latent variable $z$ enables multimodal predictions by going into the LSTM decoder with the last hidden features $h_{obs}$ during observation.  During the forecasting phase, the predicted motions $\hat{M}^{i}_{t_{obs}:t_{obs+pred}}$ are sequentially generated in the decoder.
	
	\subsection{Model Training}
	In order to encourage the coherence of temporal-neighboring features, we utilize a regularization loss  $L_c$ inspired by \cite{synnaeve2016temporal}, which is defined as follows:
	\begin{equation}
	L_{c} = \left\{\begin{matrix}
	1-\cos (h_{t_1}^i,h_{t_2}^i),\quad\quad\quad\quad|t_1-t_2|<q \\
	\max(0,\ \cos(h_{t_1}^i,h_{t_2}^i)-\textrm{margin}),otherwise\\
	\end{matrix}\right.
	\label{eq7-2}
	\end{equation}
	where $cos$ is cosine similarity and $margin$ is a hyperparameter. The pair-wise features $(h_{t_1}^i,h_{t_2}^i)$ are randomly sampled in a batch.  $L_c$ maximizes the similarity of features within a queue length (where frames are likely to strongly correlated with each other), while penalizing the similarity of features over a queue length (where frames are likely to belong to different motion patterns). The total loss function combines the regularization term $L_c$ and the variety loss (the second term) followed by \cite{gupta2018social} as:
	\begin{equation}
	Loss= \lambda L_c + \min_{m}\left \| M^i_{t_{obs}:t_{obs+pred}} - \hat{M}^{i(m)}_{t_{obs}:t_{obs+pred}} \right \|_2 ,
	\label{eq14}
	\end{equation}
	where $\lambda$ is a trade-off parameter. The variety loss computes the $L2$ distance between the best of $m$ predictions and the ground truth, which encourages to cover the space of outputs that conform to the past trajectory.

	\section{Experiments}
	\subsection{Datasets and Evaluation Metrics}
	We evaluate our method on three datasets (ETH \cite{pellegrini2009you}, UCY \cite{lerner2007crowds}, SDD \cite{robicquet2016learning}). ETH contains two subsets named ETH and HOTEL. UCY consists of three subsets, called UNIV, ZARA1 and ZARA2. Totally, there are 1,536 trajectories of pedestrian in the crowd collected in 5 scenes. We observe for 3.2s (8 frames) and the predict the motions the next 4.8s (12 frames) for every pedestrian simultaneously. For the data split and evaluation, we follow the leave-one-out method in \cite{gupta2018social}. SDD dataset has large volume with complex scenes. It contains 60 bird-eye-view videos with corresponding trajectories that involves multiple kinds of agents (pedestrian, bicyclist, .etc). The observation duration is 3.2s and the prediction duration ranged from 1s to 4s. We divide the dataset into 16,000 video clips and follow a 5-fold cross-validation setup.
	
	Commonly used Euclidean-based metrics like ADE and FDE neglect the temporal correlation of motion patterns. An illustration example is shown in Fig.\ref{figure3b}.  In order to supplement this loophole, we introduce a new metric that requires no assumptions about the temporal distribution of trajectories, Temporal Correlation Coefficient (TCC). The TCC is defined as:
	\begin{equation}
	TCC =\frac{1}{2} (TCC_x + TCC_y).
	\label{eq15}
	\end{equation}
	\begin{equation}
	TCC_{x} = \frac{1}{N} \sum_i^N \frac{Cov(\hat{x^i}, x^i)}{\sqrt{D(\hat{x^i})D(x^i)}}, \quad\quad
	TCC_{y} = \frac{1}{N} \sum_i^N \frac{Cov(\hat{y^i}, y^i)}{\sqrt{D(\hat{y^i})D(y^i)}},
	\label{eq16}
	\end{equation}
	where the ground truth trajectory  for the agent $i$ is $ M^i=( x^i,  y^i)$. The corresponding predictions are denoted as $\hat{M^i}=( \hat{x^i},  \hat{y^i} )$.  From the equations above, we can observe that TCC ranges from $\left [ -1,1 \right ]$. A high TCC implies the predictions capture the time-varying motion patterns greatly, whereas a negative TCC denotes a weak capture of temporal correlation. 
	
	For the evaluation, the metric ADE (Average Distance Error) denotes  the average L2 distance between the predictions and ground truth, and the metric FDE (Final Distance Error) reflects the L2 distance between the predictions and ground truth in the final frame. The TCC (Temporal Correlation Coefficient) is used to evaluation the temporal correlation of motion pattern in predictions.
	
	\subsection{Implementation Details}
	We preprocess the input motion state as the relative position. The size of hidden feature and the dimension of latent variable are set as 32 and 16 respectively. The convolutional part for scene is three-layer with kernel size as 10, 10, 1. The subsequent FC layer is a 16 $\times$ 16  transformation with sigmoid activation. The queue length is set as 4, 2, 3 for the ETH dataset, ZARA datasets and otherwise datasets respectively. For the loss function, the $\lambda$ and $margin$ for the regularizer $L_c$ are 0.1 and 0.5 respectively. The $m$ in the variety loss is set as 20. The batch size is 64 and the learning rate is 0.001 with Adam optimizer.

	\subsection{Standard Evaluations}
	We choose two basic methods linear models and LSTM \cite{hochreiter1997long}, and several representative state-of-the-art methods for comparison. S-LSTM \cite{alahi2016social} and SGAN \cite{gupta2018social} are the famous deterministic method
	and stochastic method that combine deep learning with spatial-only interaction respectively. The most recent approaches like Sophie \cite{sadeghian2019sophie}, MATF \cite{zhao2019multi} and STGAT \cite{huang2019stgat} incorporate information from either static scenes or spatio-temporal dependencies. To verify the effectiveness of Individual Context Module and Social-aware Context Module, we adopt two variant of our methods, Ours$_{\textrm{IC}}$ and Ours$_{\textrm{IC-SC}}$. In accord with \cite{gupta2018social,sadeghian2019sophie,zhao2019multi,huang2019stgat}, the latent variables employed in the variants Ours$_{\textrm{IC}}$ and Ours$_{\textrm{IC-SC}}$ are sampled from $\mathcal{N}(0,1)$,  and the results are reported by sampling 20 times to choose the best prediction. $``$Rel. gain$"$ shows the relative ADE gain of our full model (marked in red) compared with the latest method STGAT (marked in blue).
	\setlength{\tabcolsep}{3.2pt}
	\begin{table}[t]
		\begin{center}
			\caption{Quantitative comparisons on the ETH and UCY datasets. ADE/FDE are reported as in meters. All the models observe for 3.2 seconds and  predict for the next 4.8 seconds.}
			\label{table1}
			\begin{tabular}{l|lllll|l}
				\hline
				Methods  & ETH & HOTEL & UNIV & ZARA1 & ZARA2   &  Avg.\\
				\hline
				Linear & 0.91/1.97 & 0.42/0.81 & 0.70/1.33 & 0.58/1.23 &0.64/1.21  & 0.65/1.31 \\
				LSTM & 1.16/2.20 & 0.48/0.86& 0.57/1.20& 0.47/0.99&0.39/0.82 & 0.61/1.21 \\			
				S-LSTM & 0.84/1.85 & 0.45/0.86 & 0.55/1.14 & 0.35/0.76 & 0.36/0.77 & 0.51/1.08 \\
				SGAN & 0.77/1.41 & 0.44/0.88 & 0.75/1.50 & 0.35/0.69 & 0.36/0.73 & 0.53/1.04 \\
				Sophie & 0.70/1.43 & 0.76/1.67 & 0.54/1.24 & 0.30/0.63 & 0.38/0.78 & 0.54/1.15 \\
				MATF & 1.01/1.75 & 0.43/0.80& 0.44/0.91  & 0.26/0.45  &  0.26/0.57  & 0.48/0.90             \\
				\textcolor{blue}{STGAT} & 0.76/1.61 & 0.32/0.56 &0.52/1.10 & 0.34/0.74& 0.31/0.71&  0.45/0.94 \\
				\hline
				Ours$_{\textrm{IC}}$ & 0.70/1.35 & 0.30/0.52 & 0.52/1.09 & 0.37/0.76 & 0.30/0.63 & 0.44/0.87 \\
				Ours$_{\textrm{IC-SC}}$ & 0.70/1.36 & 0.27/0.48 & 0.51/1.08 &  0.35/0.74 &  0.28/0.61 & 0.42/0.85  \\
				\textcolor{red}{Ours (full)} & \textbf{0.66/1.21} & \textbf{0.27/0.46} & \textbf{0.50/1.07} & \textbf{0.33/0.68} &  \textbf{0.28/0.60} & \textbf{0.41/0.80} \\
				\hline
				Rel. gain & +13.15\% & +15.63\% & +3.85\% & +2.94\%  & +9.68\% & +9.05\% \\ 					
				\hline
			\end{tabular}
		\end{center}
		\vskip -0.5cm
	\end{table}
	\setlength{\tabcolsep}{7pt}
	\begin{table}[ht]
		\begin{center}
			\caption{Quantitative comparisons on the SDD dataset.  ADE/FDE are reported in pixel coordinates at 1/5 resolution followed \cite{lee2017desire}. All the models observe the agents 3.2 seconds and then make predictions in the next 1$\sim$4 seconds.} 
			\label{table2}
			\begin{tabular}{l|llll|l}
				\hline
				Methods  & 1.0 sec & 2.0 sec & 3.0 sec & 4.0 sec  &  Avg.\\
				\hline
				Linear & 1.52/2.90& 2.10/4.32& 3.01/6.23& 3.10/6.33& 2.43/4.95 \\			
				LSTM & 1.38/2.05 & 2.04/3.48 &3.02/5.91  & 3.07/5.96&2.38/4.35  \\
				S-LSTM & 1.33/2.02 &  2.00/3.46&  3.03/5.86& 3.05/5.91& 2.35/4.31\\			
				SGAN & 1.37/2.26 & 2.50/4.95 & 2.82/5.54 & 2.85/5.78 & 2.39/4.63 \\
				\textcolor{blue}{STGAT} & 1.19/1.68 &  1.69/2.90 & 2.70/5.22 & 2.83/5.37 & 2.10/3.79 \\
				\hline
				Ours$_{\textrm{IC}}$ & 1.10/1.66 & 1.70/2.90 & 2.55/5.02 & 2.65/5.13 & 2.00/3.68 \\
				Ours$_{\textrm{IC-SC}}$ & 1.09/1.65 & 1.68/2.87 & 2.52/4.96 & 2.61/5.08 & 1.98/3.64  \\
				\textcolor{red}{Ours (full)} & \textbf{1.08/1.63} &  \textbf{1.64/2.83}&  \textbf{2.48/4.91} &  \textbf{2.57/5.02} &  \textbf{1.94/3.60}  \\ 	
				\hline
				Rel. gain &  +10.19\%&  +2.96\% &  +8.14\% & +9.19\% & +7.62\% \\ 					
				\hline			
			\end{tabular}
		\end{center}
	\end{table}
	\setlength{\tabcolsep}{1.4pt} 
	
	As presented in the Table \ref{table1} and \ref{table2}, linear method and LSTM suffer from bad performance since they are too shallow to consider the surrounding context. Compared with the variant Ours$_{\textrm{IC}}$ with the state-of-the-methods, Ours$_{\textrm{IC}}$ has already shown advantages across different datasets. It indicates us that the explicit temporal dependencies extraction among multiple frames is valuable to enhance the performance. Ours$_{\textrm{IC-SC}}$ makes some improvement with social-aware features refinement. The performance gap between Ours$_{\textrm{IC-SC}}$  and our full model empirically shows that the guidance of static scene context is useful for multimodal predictions. 
	
	\section{Discussion}
	\subsection{Memory Cell Visualization} 
\begin{figure*}[t]
		\centering
		\subfigure{
			\begin{minipage}[t]{0.48\linewidth}
				\centering
				\includegraphics[width=1.8in, height=1.2in]{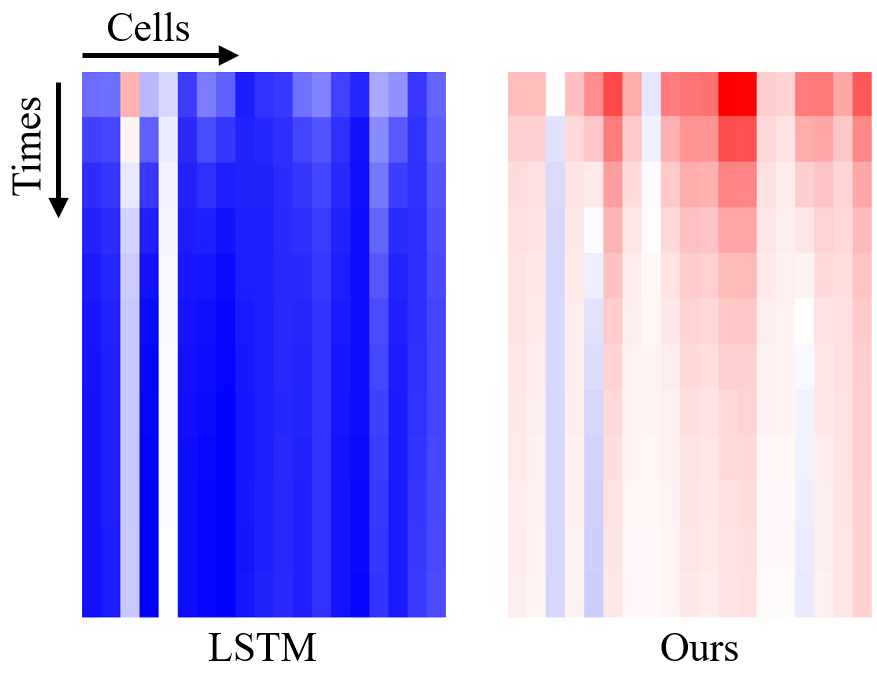}
				\label{figure4a}
			\end{minipage}	
		}
		\subfigure{
			\begin{minipage}[t]{0.48\linewidth}
				\centering
				\includegraphics[width=1.8in, height=1.2in]{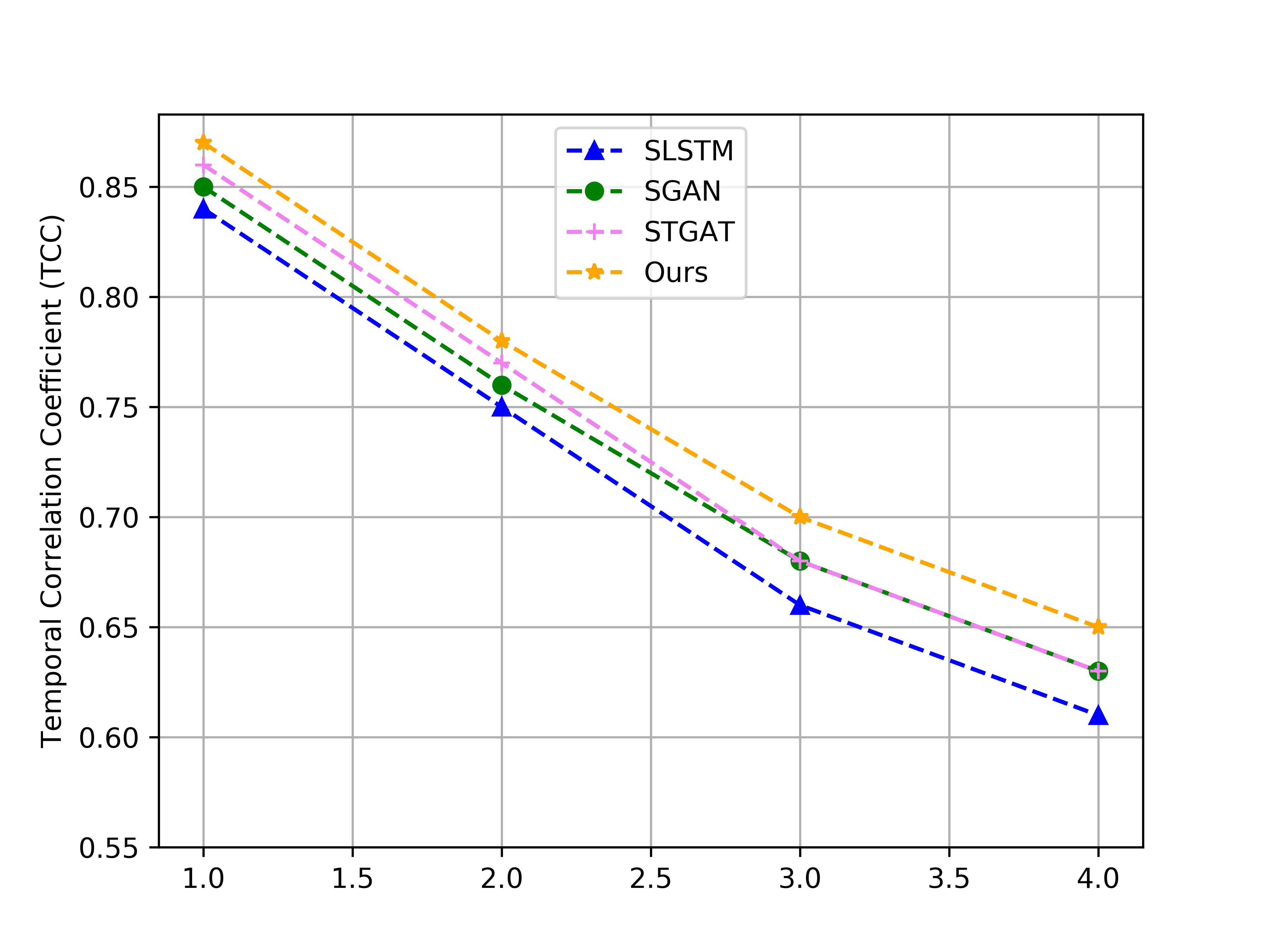}
				\label{figure4b}
			\end{minipage}%
		}
		\centering
		\caption{\textbf{(a):} Memory cell visualization for LSTM and our method. Although the memory capacity  decreases over time for both models (light blue $ \rightarrow$ dark blue for LSTM, dark red $ \rightarrow$  light red for ours), most of the cells in ours remain positive, which implies that they keep track of the evolving motion patterns from the historical context. \textbf{(b):} Comparison of TCC among state-of-the-art methods and ours. Our method enjoy high TCC, which indicates an effective capture of temporal correlation.}
		\label{figure4}
	\end{figure*}

As illustrated in Fig.\ref{figure4a}, we compare the memory capacity of LSTM and our method via cell activation. Red denotes a positive cell state, and blue denotes a negative one. The most of cells in  vanilla LSTM are  negative. In contrast, our model keeps track of the context throughout the prediction. Although the memory capacity of our model recedes over time (from dark red to light red), it still stays active. These results inspire us that the frame-by-frame observation used by LSTM is prone to get short sight. Instead, explicit modeling on the dependencies in multiple frames improves the captures of long-term motion.
	
	\subsection{The Capture of Motion Pattern} Fig.\ref{figure4b} summarizes the quantitative results of the TCC for different methods in the SDD dataset. By learning the spatio-temporal context of dynamic agents, Our method outperforms the state-of-the-art methods (SLSTM, SGAN, STGAT), especially on the long-term predictions(4s). TCC declines consistently for different methods as the prediction duration goes on. It is reasonable since the temporal correlation of longer trajectory is harder to be learned.
	
	\subsection{Exploration on the Queue Length}
	An important setup in our model is the proposed queue that understand the long-term motion. Hence, we study the effect of queue length on non-linear cases which are usually treated as hard cases. As shown in the Table \ref{table3}, linear model and LSTM  suffer from the unsatisfactory performance. Compared with the methods S-LSTM, SGAN and STGAT, our model enjoys relatively lower error. With the variation of queue length, our performance is robust and competitive against state-of-the-art methods. The model with long queue length ($q=4$) is the most suitable for the ETH dataset, where most of the trajectories are highly non-linear. Short queue length ($q=2$) works better in the ZARA1 and ZARA2 datasets,  where the trajectories are faintly non-linear. Compared with linear trajectories, We speculate that the motion states of non-linear trajectory are temporally correlated within a relatively long range, where long queues capture long-term motions better than short queues.
	
	\setlength{\tabcolsep}{1.6pt}
	\begin{table}[t]
		\begin{center}
			\caption{Comparisons of non-linear trajectories on the ETH and UCY datasets. ADE/FDE are reported as in meters. ``Ours'' denotes the proposed full model.}
			\label{table3}
			\begin{tabular}{l|lllll|l}
				\hline
				Methods & ETH & HOTEL & UNIV & ZARA1 & ZARA2  & Avg.\\

				\hline

				Linear &1.01/2.21&0.50/0.96&0.80/1.56&0.62/1.33&0.99/1.95& 0.78/1.60 \\
				LSTM &1.08/2.28 & 0.57/1.02& 0.65/1.37& 0.49/1.04& 0.60/1.30& 0.68/1.40\\			
				S-LSTM &  0.92/2.01 & 0.48/0.92 & 0.61/1.29 &  0.38/0.79 & 0.50/1.06 & 0.58/1.21  \\
				SGAN & 0.86/1.54 & 0.52/1.03 & 0.81/1.61 & 0.39/0.76 & 0.49/0.97 & 0.61/1.18  \\
				STGAT & 0.86/1.67  &  0.39/0.71 & 0.60/1.33 & 0.39/0.85 & 0.50/1.14 & 0.55/1.14 \\	
				\hline
				Ours (q=2) &0.73/1.37  & 0.34/0.63 & 0.61/1.29 &   \textbf{0.36/0.76}& \textbf{0.48/1.04} & 0.50/1.02 \\
				Ours (q=3) & 0.71/1.35 & \textbf{0.33/0.58} &  \textbf{0.60/1.28} & 0.38/0.81  & 0.49/1.09 & \textbf{0.50/1.01} \\					
				Ours (q=4) & \textbf{0.70/1.29} & 0.34/0.60  & 0.60/1.31  &  0.39/0.83  &  0.49/1.09  & 0.50/1.02   \\			
				\hline
			\end{tabular}
		\end{center}
	\end{table}
	\setlength{\tabcolsep}{0.5pt}
	
	\subsection{Social Behaviors Understanding}
\begin{figure*}[t]
		\centering
		\begin{minipage}[t]{1\linewidth}
			\centering
			\includegraphics[width=4.5in, height=0.3in]{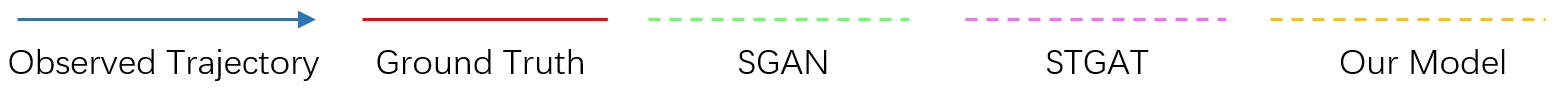}
			\label{figure5z}
		\end{minipage}%
		
		\subfigure[Walk in parallel]{
			\begin{minipage}[t]{0.33\linewidth}
				\centering
				\includegraphics[width=1.4in, height=2.2in]{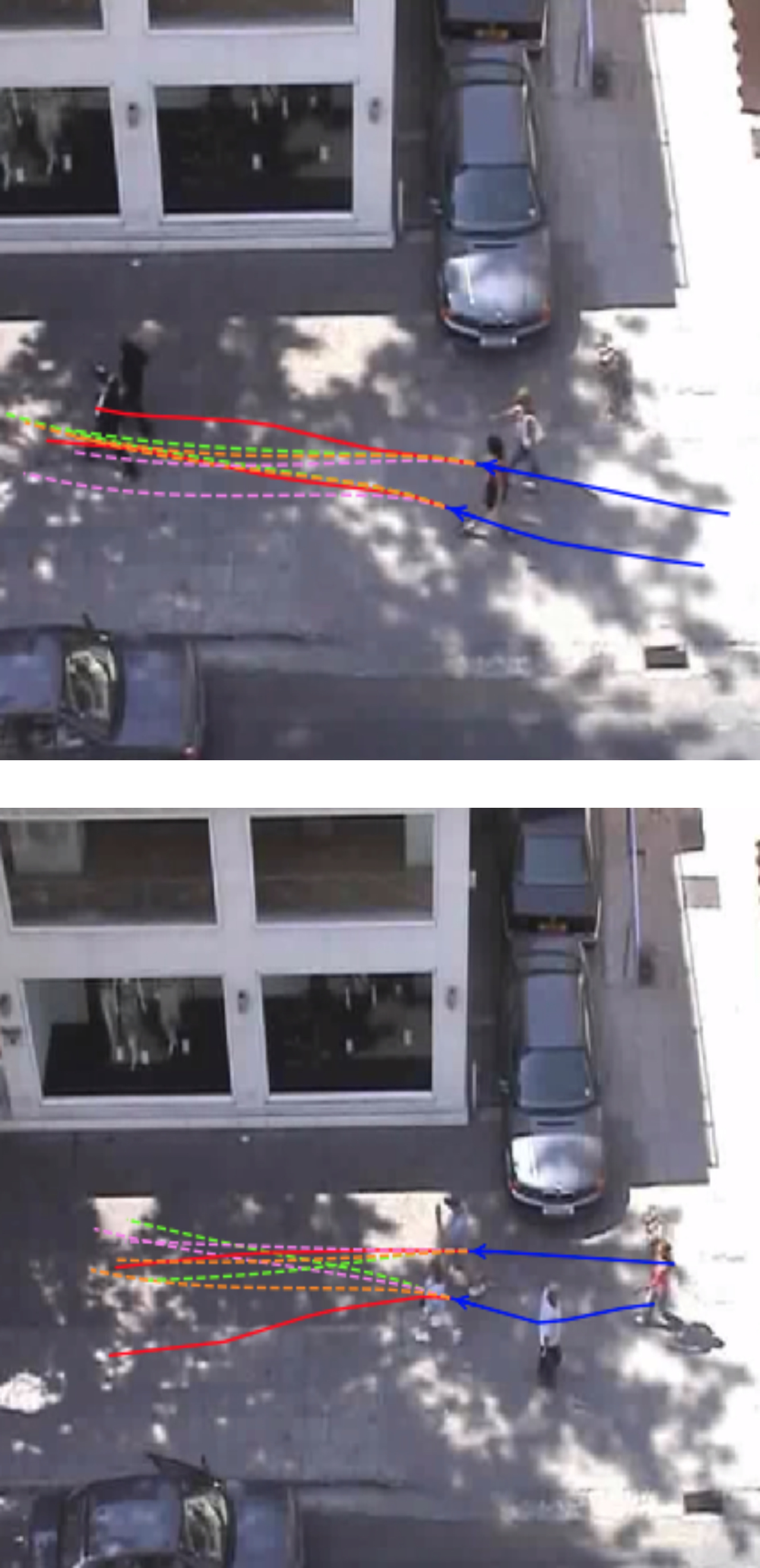}
				\label{figure5a}
			\end{minipage}%
		}%
		\subfigure[Turning]{
			\begin{minipage}[t]{0.33\linewidth}
				\centering
				\includegraphics[width=1.4in, height=2.2in]{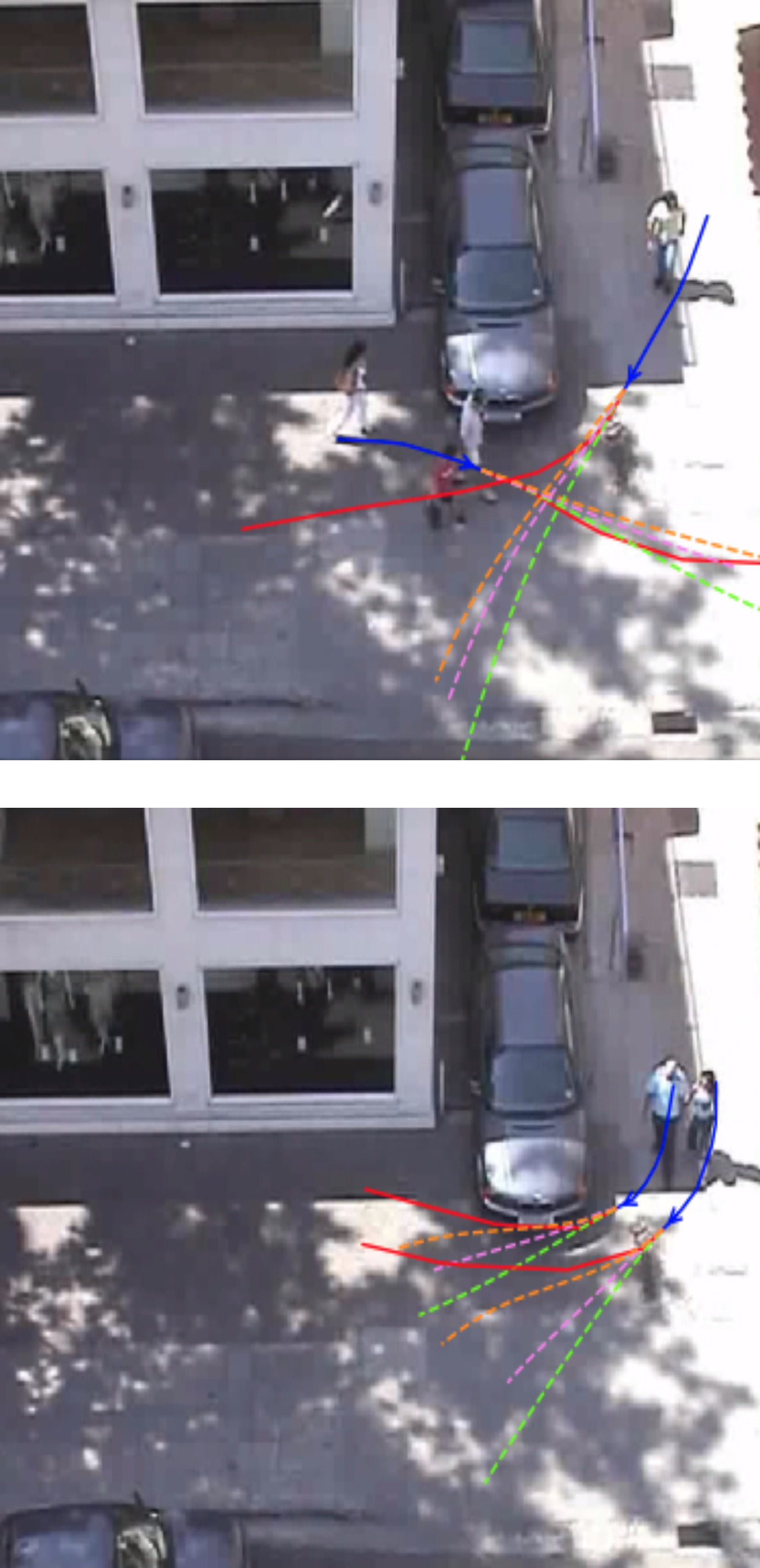}
				\label{figure5b}
			\end{minipage}%
		}%
		\subfigure[Face-to-face]{
			\begin{minipage}[t]{0.33\linewidth}
				\centering
				\includegraphics[width=1.4in, height=2.2in]{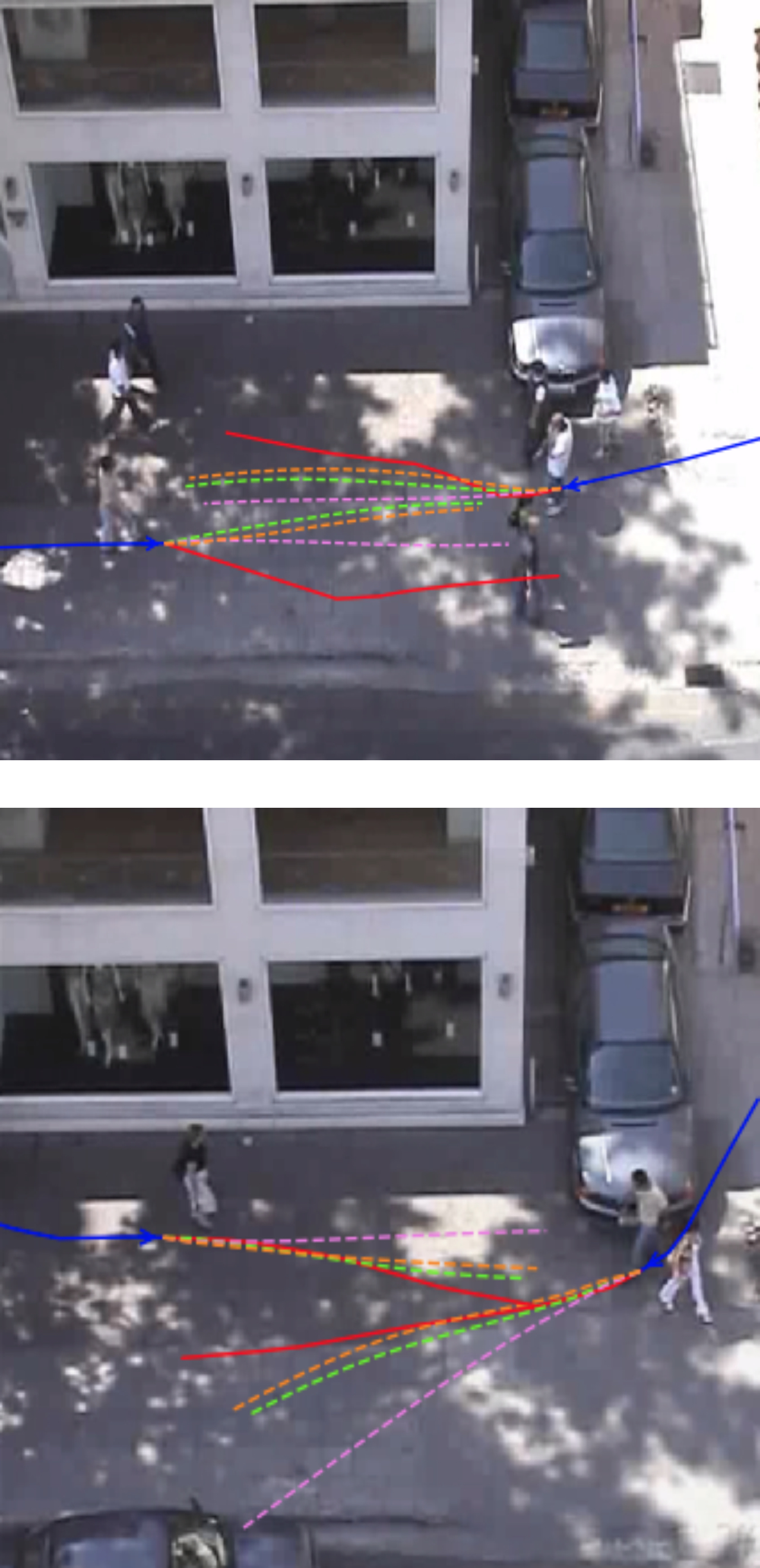}
				\label{figure5c}
			\end{minipage}%
		}%
		\centering
		\caption{Comparison among our model, SGAN and STGAT in different scenarios. These visualizations show that 1): Our model is capable of generating convincing trajectories that are closer to the ground truth than other state-of-the-art methods.  2): The social interactions across agents are well captured by our model. Our predictions avoid collision during the whole forecasting period.}
		\label{figure5}
	\end{figure*}

	In the scenario of real-world applications, it is imperative to handle the social interactions in the multi-agent system. Therefore, we verify whether our method well perceive the social behaviors in the crowd. As shown in the Fig.\ref{figure5}, we select three common scenarios involve social behaviors, ``Walk in parallel'', ``Turing'' and ``Face-to-Face''. From the comparison between the ground truth and various predictions, we could observe that the trajectories predicted by our method are close with ground truth. Moreover, our predictions are reliable that no collisions or large deviations appear during the whole forecasting period. It indicates that our model predict multiple trajectories for	each agent that cohere in time and space with the other agents.

	\subsection{Analysis of Multimodal Predictions}
\begin{figure*}[t]
		\centering
		
		\begin{minipage}[t]{1\linewidth}
			\centering
			\includegraphics[width=4.5in, height=0.3in]{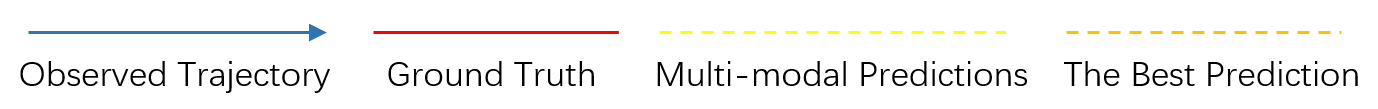}
			
			\label{figure6z}
		\end{minipage}%
		
		\subfigure[Crossroad]{
			\begin{minipage}[t]{0.24\linewidth}
				\centering
				\includegraphics[width=1.0in, height=2.0in]{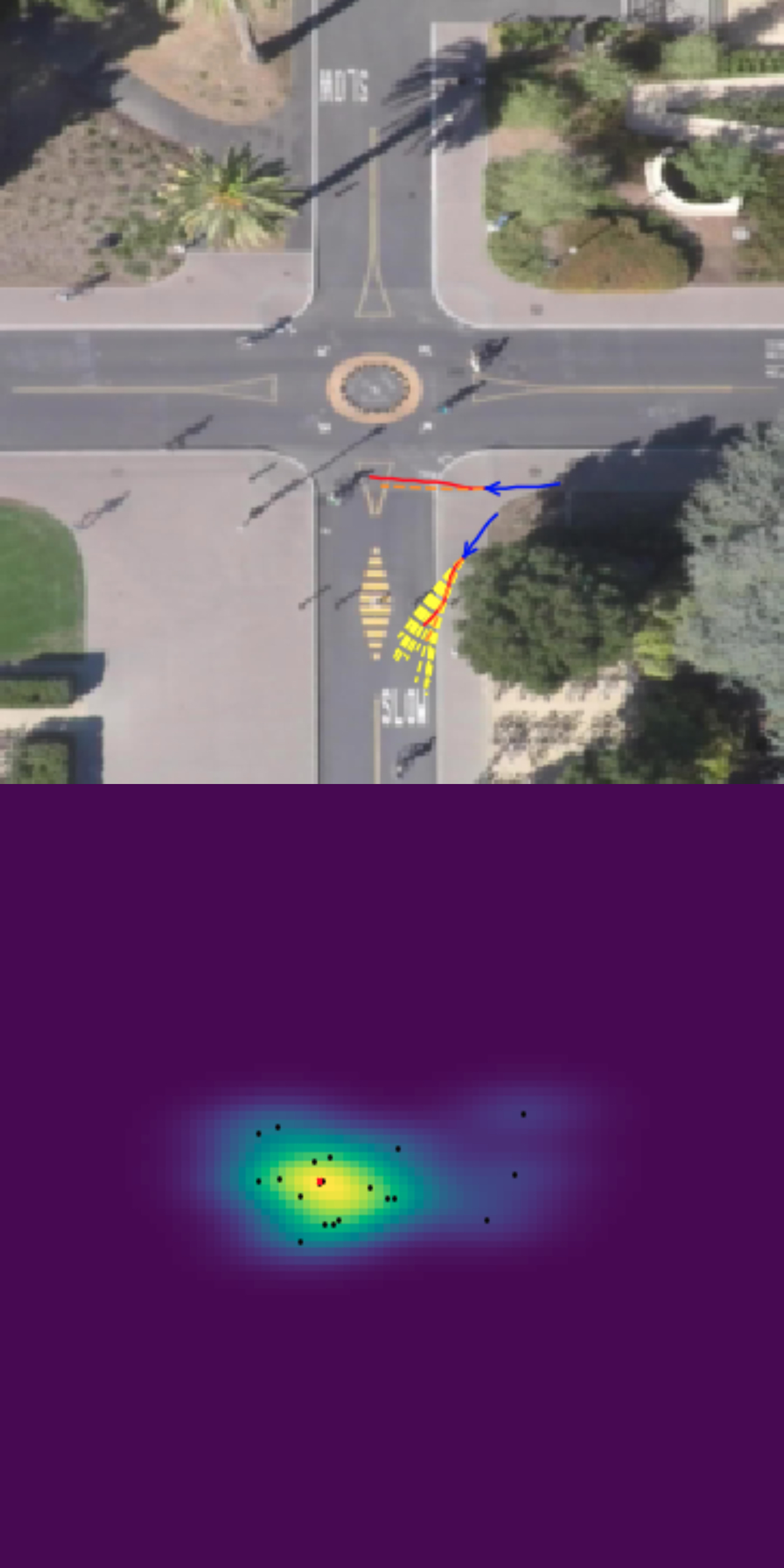}
				\label{figure6a}
			\end{minipage}%
		}%
		\subfigure[Intersection]{
			\begin{minipage}[t]{0.24\linewidth}
				\centering
				\includegraphics[width=1.0in, height=2.0in]{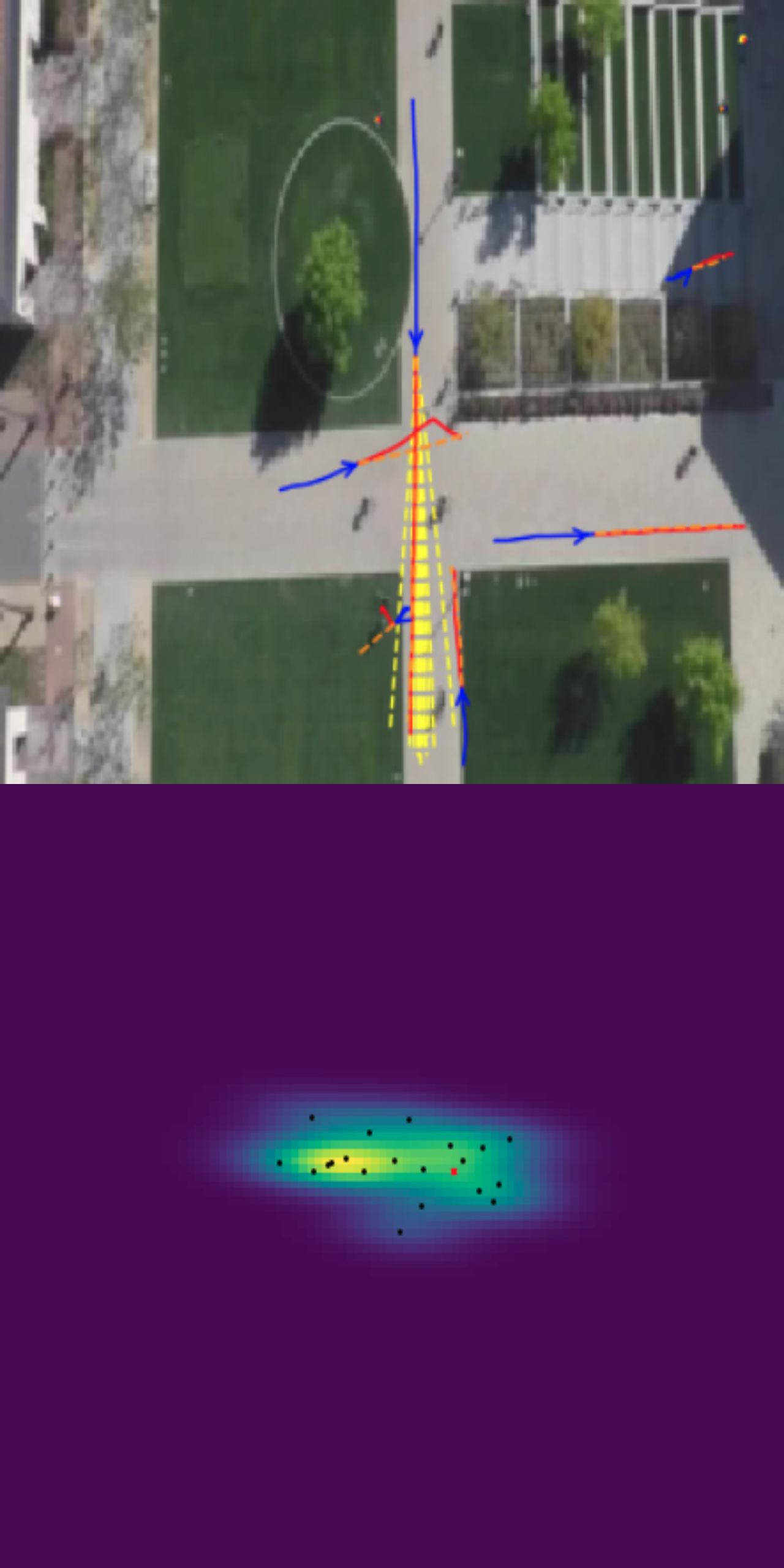}
				\label{figure6b}
			\end{minipage}%
		}%
		\subfigure[Sideway]{
			\begin{minipage}[t]{0.24\linewidth}
				\centering
				\includegraphics[width=1.0in, height=2.0in]{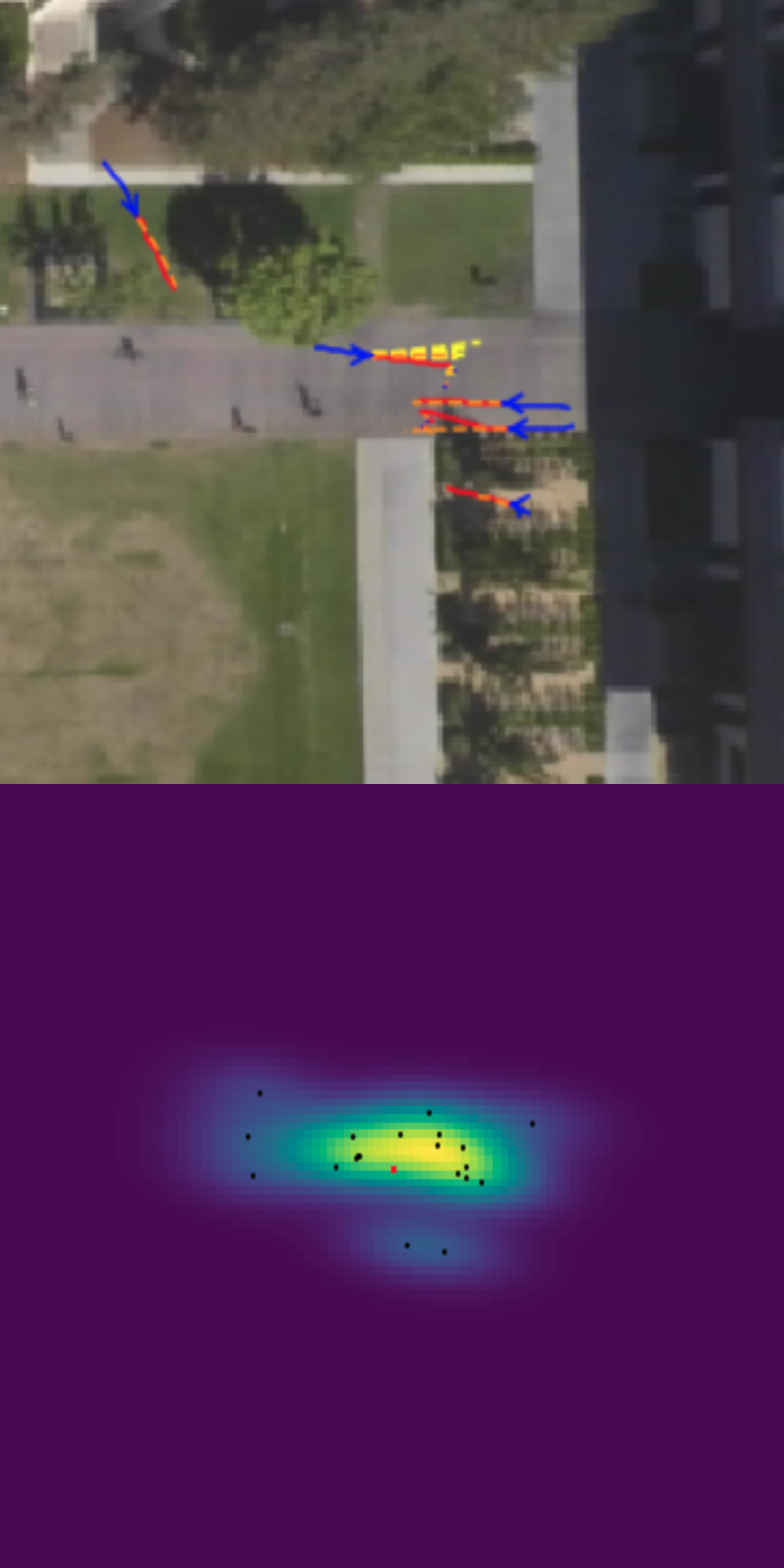}
				\label{figure6c}
			\end{minipage}
		}%
		\subfigure[Corner]{
			\begin{minipage}[t]{0.24\linewidth}
				\centering
				\includegraphics[width=1.0in, height=2.0in]{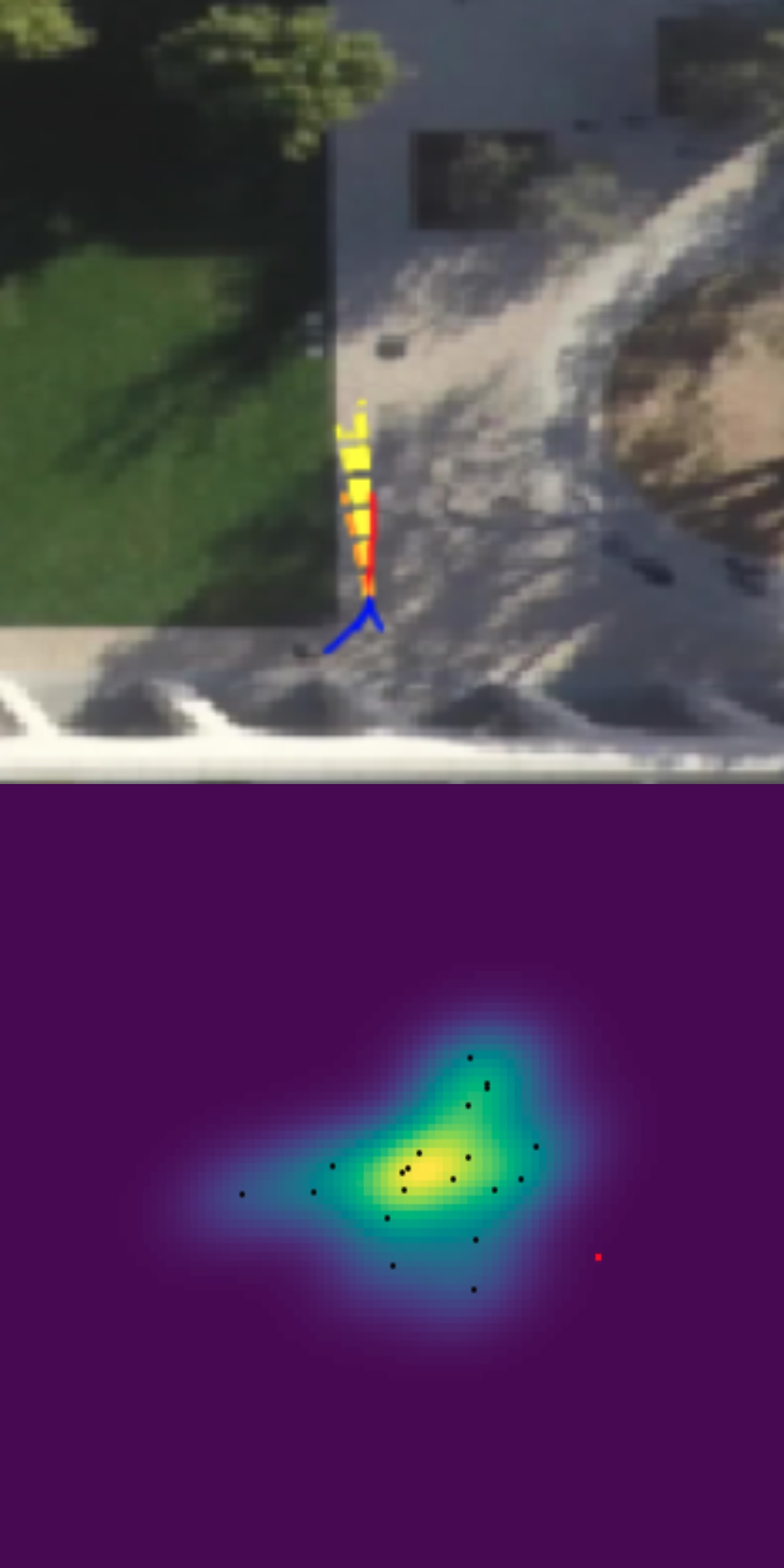}
				\label{figure6d}
			\end{minipage}
		}%
		\centering
		\caption{The visualization of the multimodal predictions in four scenes. \textbf{Top row:} we plot multiple possible future trajectories for one agent of interest.  \textbf{Bottom row:} we visualize the distribution heatmap of  destinations (location at the final frame) via kernel density estimation. The predicted destinations and ground truth are shown as black points and red point respectively. The distribution heatmap shows that our model not only provides semantically meaningful predictions, but also enjoys low uncertainty.}
		\label{figure6}
	\end{figure*}

	In order to evaluate the quality of  multimodal predictions, we visualize the diverse trajectories predicted by our model. The top row in the Fig.\ref{figure6} reports the multimodal predictions for one agent of interest. Rather than using a predefined Gaussian noise, the learnable latent variable $z$ benefits from the semantic cues of the static scene context. Our predictions (yellow lines) suggest plausible trajectories that are close to the ground truth (red line), instead of producing a wide spread of candidates randomly. For instance, in the scenario of ``Crossroad'' and ``Intersection'', it is possible for the target agent to turn left or right at the endpoint of observation. Our model provides predictions that in line with common sense. In the scenario of ``Sideway'' and ``Corner'', the target agent has limited choices for the future trajectories due to the constraint of scene layout. In these kinds of scenarios, all our predictions are moving towards reasonable directions. Hence, our model has good interpretability with the incorporation of scene information.
	
	In the bottom row, we investigate the uncertainty of predictions with distribution heatmap. Here we estimate the distribution of the predicted destination (black point) via kernel density estimation, and then apply the true destination (red point) to this distribution. The brighter the location, the more possibility that the point belongs to the distribution. Our visualization shows that  the true destination usually appear in the bright locations. It indicates our predictions enjoy low uncertainty.
	
	\section{Conclusions}
	In this paper, we proposed a novel method DSCMP to highlight the three core elements of contextual understanding, \textit{i.e.} spatial interaction, temporal coherence and scene layout, for multi-agent motion prediction.
	We designed a differentiable  queue mechanism embedded on LSTM to capture the spatial interactions across agents and temporal coherence in long-term motion.
	And a learnable latent variable was introduced to learn the semantics of scene layout.  
	In order to understand the uniqueness of DSCMP, we also proposed a metric Temporal Correlation Coefficient (TCC) to evaluate the temporal correlation of predicted motion. 
	Extensive experiments on three benchmark datasets demonstrate the effectiveness of our proposed method. 
	For the future research on autonomous applications, this work sheds a light on the modelling of spatio-temporal dependencies in multiple frames and the semantic cues from scene layout.

	\section*{Acknowledgments}
	This work is partially supported by the SenseTime Donation for Research, HKU Seed Fund for Basic Research, Startup Fund and General Research Fund No.27208720. This work is also partially supported by the Major Project for New Generation of AI under Grant No. 2018AAA0100400 and by the National Natural Science Foundation of China under Grant No. 61772118. 
	
	\clearpage
	%
	%
	\bibliographystyle{splncs04}
	\bibliography{egbib}
\end{document}